# The Geometry of Culture:
# Analyzing Meaning through Word Embeddings

Austin C. Kozlowski[1]
Matt Taddy[2,3]
James A. Evans[1]

[1] University of Chicago, Department of Sociology
[2] University of Chicago, Booth School of Business
[3] Amazon

We demonstrate the utility of a new methodological tool, neural-network word embedding models, for large-scale text analysis, revealing how these models produce richer insights into cultural associations and categories than possible with prior methods. Word embeddings represent semantic relations between words as geometric relationships between vectors in a high-dimensional space, operationalizing a relational model of meaning consistent with contemporary theories of identity and culture. We show that dimensions induced by word differences (e.g., *man – woman*, *rich – poor*, *black – white*, *liberal – conservative*) in these vector spaces closely correspond to dimensions of cultural meaning, and the projection of words onto these dimensions reflects widely shared cultural connotations when compared to surveyed responses and labeled historical data. We pilot a method for testing the stability of these associations, then demonstrate applications of word embeddings for macro-cultural investigation with a longitudinal analysis of the coevolution of gender and class associations in the United States over the 20th century and comparative analysis of historic distinctions between markers of gender and class in the U.S. and Britain. We argue that the success of these high-dimensional models motivates a move towards "high-dimensional theorizing" of meanings, identities and cultural processes.

INTRODUCTION

      A vast amount of information about what people do, know, think, and feel lies preserved in digitized text and an increasing proportion of social life now occurs natively in this medium. Available sources of digitized text are wide ranging, including collective activity on the web, social media, and instant messages as well as online transactions, medical records, and digitized

letters, pamphlets, articles, and books (Evans and Aceves 2016; Grimmer and Stewart 2013). This growing supply of text has elicited demand for natural language processing and machine-learning tools to filter, search, and translate text into valuable data. The analysis of large digitized corpora has already proven fruitful in a range of social scientific endeavors including analysis of discourse surrounding political elections and social movements, the accumulation of knowledge in the production of science, and communication and collaboration within organizations (Christopher Bail 2012; Evans and Aceves 2016; Foster, Rzhetsky, and Evans 2015; Grimmer 2009; Goldberg et al. 2016).

Although text analysis has long been a cornerstone for the study of culture, the impact of "big data" on the sociology of culture remains modest (Bail 2014). A fundamental challenge for the computational analysis of text is to simultaneously leverage the richness and complexity inherent in large corpora while producing a representation simple enough to be intuitively understandable, analytically useful and theoretically relevant. Moreover, turning text into data (Grimmer and Stewart 2013) requires credible methods for (1) evaluating the statistical significance of observed patterns and (2) disciplining the space of interpretations to avoid the tendency to creatively confirm expectations (Nickerson 1998). While past research has made strides towards overcoming these challenges in the study of culture, critics continue to argue that existing methods fail to capture the nuances of text that can be gleaned from interpretive text analysis (Biernacki 2012).

In this paper, we demonstrate the utility of a new computational approach – neural-network word embedding models – for the sociological analysis of culture. We show that word embedding models are able to capture more complex semantic relations than past modes of computational text analysis and can prove a powerful tool in the study of cultural categories and associations. Word embeddings are high-dimensional vector-space models[1] of text in which each unique word in the corpus is represented as a vector in a shared vector space (Mikolov, Yih, and Zweig 2013; Pennington, Socher, and Manning 2014). Methods similar to word embeddings, such as Latent Semantic Analysis (LSA) or Indexing (LSI), have existed in various forms since the 1970s (Dumais 2004). Recent breakthroughs in auto-encoding neural networks and advances in computational power have enabled a new class of word embedding models that incorporate relevant information about word contexts from highly local windows of surrounding words rather than an entire surrounding document. As a result, these new word embedding models distil an encyclopedic breadth of subtle and complex cultural associations from large collections of text by training the model with local word associations a human might learn through ambient exposure to the same collection of language.

In word embedding models, words are assigned a position in a vector space based on the context that word shares with other words in the corpus. Words that share many contexts are positioned near one another, while words that inhabit very different contexts locate farther apart. Previous work with word embedding models in computational linguistics and natural language processing has shown that words frequently sharing linguistic contexts, and thus located nearby in the vector space, tend to share similar meanings. However, semantic information is encoded not only in the clustering of words with similar meanings. In this paper we present evidence that the very dimensions of these vector space models closely correspond to meaningful "cultural dimensions" such as race, class, and gender. We show that the positioning of word vectors along culturally salient dimensions within the vector space captures how concepts are related to one another within cultural categories. For example, projecting occupation names on a "gender dimension," we find that traditionally feminine occupations such as "nurse" and "nanny" are positioned at one end the dimension and traditionally masculine occupations such as "engineer" and "lawyer" are positioned at the opposite end. This occurs because with each local context that "nurse" shares with feminine words like "she," "her," and "woman," it is nudged towards the feminine pole of the gender dimension, while each time "engineer" shares a context with terms like "his," "him," and "man," it is nudged toward the masculine pole.

---

[1] Word embedding models are sometimes considered and referred to as "low dimension" techniques relative to the number of words used in text (e.g., 20,000) because they reduce this *very* high dimensional word space. Nevertheless, considered from the perspective of one, two or three dimensional models common in the analysis of culture, these spaces are much more complex, and reproduce much more accurate total associations, as shown below.

We emphasize that word embedding models not only grant insight into the semantic structure of a given cultural system, but can also be productively used to analyze cultural difference and change. By comparing word embedding models trained on texts produced in different cultural contexts, it is possible to directly examine differences in the in the meanings of terms and categories between social groups. Similarly, comparing word embedding models trained on texts from sequential time periods make it possible to investigate how cultural associations shift historically.

In the following analyses, we utilize word embedding models trained on contemporary texts, texts stretching back over a century, and texts from distinct cultures to demonstrate the broad potential of word embedding models for cultural analysis. The following analyses proceed in four steps. First, we compare semantic relations derived from word embedding models to results from survey data measuring cultural associations in order to demonstrate the ecological validity of word embedding models in capturing widespread associations with race, class, and gender. Second, we analyze word embedding models trained on texts published in the United States spanning the entire 20th century to investigate macro-historic cultural trends, discovering slow but persistent changes in the interrelationship of the categories of gender and class. Third, we conduct a cross-national analysis, comparing text from the United States and Great Britain from the turn of the 20th century to identify subtle differences in the markers of class and gender between these two social contexts. The findings presented here demonstrate the broad utility of word embedding models for cultural analysis and motivate further development and application of these models to a diverse array of questions in sociology of culture.

FORMAL TEXT ANALYSIS IN THE STUDY OF CULTURE

Language has long held a central position in the study of culture. Cultural scholars from sociology, anthropology, and socio-linguistics have commonly understood a group's language to be a reflection of its cultural system (Whorf 1956; Levi-strauss 1963), and thus text has served as a key source of data for scholars investigating cultural categories and meaning structures. Historically, analysis of text in sociology has been dominated by qualitative, interpretive approaches, the two most common being interpretivist close-reading and systematic qualitative coding. Interpretivist text analysis, in which the researcher draws insights from a holistic deep reading of text, has done much to advance sociological understandings of culture, but suffers from clear limitations in reproducibility (Ricoeur 1981). The method of qualitative coding, in which the researcher selects a number of themes and systematically tracks their deployment in text (Glaser and Strauss 1967) can be more reproducible than a singular close reading, but still suffers from low inter-coder reliability when themes are complex or subtle. Because these dominant techniques for the analysis of culture are not easily replicable and rely upon the analyst's intuition and finesse, the study of culture in sociology has largely remained a "virtuoso affair" (DiMaggio 1997). Furthermore, both interpretivist text analysis and qualitative coding are limited by the pace of human reading, so neither are well suited for the analysis of very large corpora or entire socio-cultural domains.

  Limitations of qualitative textual analysis have motivated scholars of culture in the social sciences and humanities to develop an array of formal and quantitative methods of text analysis (Mohr 1998; D'Andrade 1995). Two such methods that have gained relative popularity in recent years are semantic network analysis and topic modeling. Semantic networks are typically constructed by treating words as nodes of a network and textual co-occurrences as links (Kaufer and Carley 1993; Carley 1994; Lee and Martin 2015; Hoffman et al. forthcoming). Examining structural characteristics of a semantic network such as words that are highly central or that bridge semantic (Corman 2002) or cultural holes (Pachucki and Breiger 2010) can provide insight into the relationships between individual words as well as the overall conceptual structure undergirding a text. Alternatively, topic modeling comprises a more recent approach that uses a well-formed probability model to enable inductive discovery of "topics" underlying a corpus, each learned as a sparse distribution over words that tend to co-occur in text (Mohr and Bogdanov 2013; Blei, Ng, and Jordan 2003). Topic modeling can detect polysemy by tracing words that exist in multiple topics, and heteroglossia, the multiple voices of a single text, by tracing the mixture of distinct topics across documents (DiMaggio, Nag, and Blei 2013; Blei 2012).

Both methods can generate important insights into the cultural system that produced a text, but there remain many sociologically and culturally important questions for which these methods are poorly suited. When corpora grow sufficiently large, standard semantic network analysis metrics fail to distinguish between concepts that are close or distant by considering topological information alone. Such networks could be made dense and their links weighted, encoding a myriad word collocations, but analysis of the resulting hairball would require a calculus that deviates widely from standard network analysis, such as one based on random walks (Shi, Foster, and Evans 2015; Rosvall and Bergstrom 2008) or implied curvature (Jost and Liu 2014). Topic modeling is also focused on creating sparse, interpretable descriptions by setting most word-topic and topic-document loadings to approximately zero. By design, this distorts optimal geometric distances between words in order to make topics human readable. As such, both networks and topic models are ill-suited for representing the multifarious associations and cultural valances that characterize *all* words in a corpus. Questions regarding how masculine or feminine, good or bad, high or low-status a given object is within a cultural system remain difficult to answer using existing formal methods of text analysis. Furthermore, investigation into the relations between cultural categories – for instance, how closely a culture's good/bad distinction relates to its masculine/feminine distinction – is also beyond the scope of prior methods.

WORD EMBEDDING MODELS AND COMPLEX SEMANTIC RELATIONSHIPS

Recent work in natural language processing has made great strides by representing word relationships in a corpus not as networks or topical clusters but as vectors in a dense, continuous, high-dimensional space (Mikolov, Yih, and Zweig 2013; Pennington, Socher, and Manning 2014; Joulin et al. 2016). These vector space models, known collectively as word embeddings, have attracted widespread interest among computer scientists and computational linguists due to their ability to capture complex semantic relations between words.

_______________________________

Figure 1 about here

_______________________________

In a word embedding model, word vectors are positioned such that words sharing similar contexts in the text will be positioned nearby in the vector space, whereas words that appear only in different and disconnected contexts will be positioned further apart. Figure 1 schematically illustrates the structure of the descriptive problem that word embeddings attempt to solve: how to represent all words from a corpus within the $k$-dimensional space that best preserves distances between $n$ words across $m$ local contexts. The solution, which we illustrate in subsequent figures, is a $n$-by-$k$ matrix of values, where $k \ll m$, bolded here where $k = 3$. An early approach to word embeddings, titled Latent Semantic Analysis (LSA) used singular-value decomposition (SVD) to discover this word matrix where the first principal component explained the most variation in the original $n$-by-$m$ word-context matrix, the second component the second most, and so on such that $k$ was typically trimmed when the marginal $k$th singular value explained arbitrarily little variation in the word-context matrix. By contrast, neural word embeddings evenly spread explanatory power across the full $k$ dimensions through heuristic optimization of a neural network with at least one "hidden-layer" of $k$ internal, dependent variables.[2] While $k \ll m$ in these models, substantial natural language corpora require $k \geq 300$ to minimize the error of word-context matrix reconstruction (Mikolov, Yih, and Zweig 2013). Although we do not make use of them in this analysis, the *m* contexts by *k* dimensions matrix in Figure 1 also retains a great deal of semantic information and has been used in concert with word embeddings to identify words that are complements versus substitutes in text (Nalisnick et al. 2016; Ruiz, Athey, and Blei 2017). It is important to note that, because the optimal distance of two vectors is a function of shared context rather than co-occurrence, words need not co-occur in order for their vectors to be positioned close together. For example, if "doctor" and "lawyer" both appear near the word "work" or "job" then the vectors for "doctor" and "lawyer" would be located near each other in the embedding, even if they never appear together in text.

---
[2] Scientists have attempted to perform these parametrically, as with exponential family embedding models, but their performance has not yet approached that of autoencoders (Rudolph et al 2016).

In word embeddings, each word in the corpus is represented geometrically as a vector and ascribed a location in the high-dimensional vector space. Distance between words in this space is typically assessed using the cosine of the angle between word vectors, rather than the Euclidean (straight-line) distance because of special properties associated with high-dimensional spaces that violate intuitions formed in two or three dimensions. For example, as the dimension of a hypersphere grows, its volume shrinks relative to its surface area as more and more of its volume resides near its surface.[3] We normalize all word vectors (Levy, Goldberg, and Dagan 2015), and therefore they lie on the surface of a hypersphere of same dimensionality as the space, and we represent proximity in terms of the angle between vectors, which is proportional to their distances on the hypersphere surface.

---

[3] Even in 3 dimensions, the surface area of a unit circle surpasses its volume.

*Word2vec*, the most widely used word embedding algorithm and the primary approach we use in the following analyses, uses a shallow, two-layered neural network architecture that optimizes the prediction of words based on shared context with other words. *Word2vec* can operate under two distinct model architectures: continuous bag-of-words (CBOW) or skip-gram. Under the CBOW architecture, the corpus is read line-by-line in a sliding window of *k* words, with *k* determined by the analyst, and previous studies finding windows of ~8 words producing most consistent results (Le and Mikolov 2014). For each word in the corpus, the algorithm aims to maximize classification of the center word *n*, given its surrounding words within a context window of size *k*. The skip-gram architecture works similarly, except that instead of predicting a word with context, the skip-gram architecture predicts context given a word. Related approaches take into account additional information such as the "global" proximity of words within an overarching document (Pennington, Socher, and Manning 2014) or even subword letter sequences in surrounding words (Joulin et al. 2016).

Because words are located together in the embedding model if they appear in similar contexts in the corpus, abutting words in the vector space tend to share similar meanings; it is common to find that a word's nearest neighbors are either its synonyms or syntactic variants. Words then tend to share a broader region of semantic space with a host of terms having related meanings. Therefore a great deal of semantic and cultural information is available simply by examining the word vectors that surround a word of interest. Kulkarni et al. (2015) have used word embedding models in this way to trace shifts in the meaning of the word "gay" over the course of the 20$^{th}$ century, from a location in the vector space beside "cheerful" and "frolicsome" to one near "lesbian" and "bisexual." Hamilton et al (2016) have similarly used word embedding models to investigate how a word's rate of semantic change, measured as change in the word's overall position in the vector space, depends upon its frequency and polysemy, finding that words occurring with high frequency change meaning more slowly and polysemous words more rapidly.

Past work with word embedding models has also shown that semantically meaningful relations can also be found between words not directly proximate in the space. From *word2vec*'s debut in 2013, neural word embeddings have intriguingly been able to solve analogy problems by applying simple linear algebra to word vectors (Mikolov et al. 2013). For example, the analogy "man is to woman as king is to ___" can be solved with a model trained on a sufficiently large body of text by performing the arithmetic operation with the word vectors *king – man + woman*, with the resulting vector most closely approximating *queen*. *Word2vec* can achieve success rates as high as 74% (Ji et al. 2016) on a challenging analogy test comprising 20 thousand questions involving semantic comparisons ranging from currency-country (*kwanza* is to Angola as *rial* is to Iran) and male-female (man is to woman as waiter is to waitress) to syntactic comparisons involving opposites, plural nouns, comparatives, superlatives, and verb conjugations (e.g., past tense, present participle) (Mikolov et al. 2013).

This ability of word embeddings to solve analogies across many conceptual dimensions derives not only from their relatively high number of dimensions, but from the type of geometry in which they are embedded. Euclidean geometry was constructed to obey Euclid's fifth postulate, which states that within a two-dimensional plane, for any given line ℓ and point *p* not on ℓ, there exists exactly one line through *p* that does not intersect ℓ. The result is a space dimensionalized by "straight" vectors and "flat" planes such that word embeddings constructed in such a space enable direct identification of semantic dimensions on which other words meaningfully project. This discussion reveals, however, that high-dimensional Euclidean embeddings are not "model free", but use a formalism tuned to directly model the kinds of semantic dimensions we feature in this paper.

Other embedding geometries could be used to model distinct cultural patterns. Scientists and engineers geometrically embed not only corpora, however, but also graphs. Social networks, power grids and the internet, with "nodes" of highly unequal degree distribution or number of connections, are often embedded in hyperbolic spaces having constant, negative curvature (Krioukov et al. 2010; Papadopoulos et al. 2012; Cvetkovski and Crovella 2009; Alanis-Lobato, Mier, and Andrade-Navarro 2016). In a hyperbolic space, infinitely many lines may go through $p$ without intersecting $\ell$ and a central node may be close to many peripheral nodes without those nodes being close to each other. Embedding language, where words have highly unequal frequencies (Zipf 1932), in a hyperbolic geometry would make discovery of analogy and the semantic dimensions underlying much less straightforward, but could potentially enable modeling of semantic hierarchy, such as the identification of holonyms and meronyms—words constituting wholes and their parts—like *hand*, *flesh*, and *fingers*; or hypernyms—words with broad meanings under which more specific words fall—such as the relationship of *color* to *red*, *green*, and *blue* (Rei and Briscoe 2014; Handler 2014). Contemporary word embeddings may not effectively model all forms of word association, but altering hidden parameters such as the curvature of the underlying geometry, could allow them to capture distinct ones.

CULTURAL DIMENSIONS OF WORD EMBEDDINGS

In this paper, we present a novel method for applying word embedding models to the sociological analysis of culture. We demonstrate that derived dimensions of word embedding vector spaces correspond closely to "cultural dimensions," such as race, class, and gender, which individuals use in everyday life to classify and categorize objects in the world. By discovering and examining these culturally meaningful dimensions in a word embedding, it is possible for an analyst to reveal where objects are located on various cultural dimensions and to determine how these dimensions are positioned relative to one another in that space. For instance, an analyst can use a word embedding model to determine whether "opera" is considered more masculine or feminine than "jazz" by projecting the word vectors corresponding to "opera" and "jazz" onto the dimension corresponding to gender in the space (e.g., *man – woman*). Similarly, the researcher can determine if "jazz" is more upper-class or more working-class than "opera" by projecting these words onto the identity dimension corresponding to class in the same space. This "dimensional" approach emphasizes that semantic meaning is contained not only in the distance between two word vectors but also in the *direction* of that difference.

The technique we present for discovery of cultural dimensions in a word embedding vector space builds upon logic for solving analogies with word embeddings. One interpretation for why *king + woman – man* ≈ *queen* in word embedding models is because (*woman – man*) corresponds closely corresponds to a "gender dimension." Adding (*woman – man*) to *king* therefore has the effect of starting at *king* and taking one step on the gender dimension in the direction of femininity. Conversely, (*man – woman*) corresponds to "one step" in the direction of masculinity on the same gender dimension. Thus, *queen + (man – woman)* ≈ *king*. An approximation of the gender dimension is captured not only by (*man – woman*), but also by any other pairs of words whose semantic difference correspond to that cultural dimension of interest. For example, *he – she*, *him – her*, *male – female*, or even *waiter – waitress* would approximate the same gender dimension as *man – woman*. Because we expect *man – woman*, *he – she*, *him – her*, and similar gender antonym pairs to all approximate the gender dimension of interest, we calculate a gender dimension by simply taking the arithmetic mean of a set of such pairs. Dimensions for class and race are similarly constructed, using appropriate sets of antonym pairs such as *rich – poor* or *black – white*, respectively[4]. We note that an analyst could also discover differential weights for each word pair by estimating them within a linear model that predicts surveyed cultural associations. A weighted sum necessarily improves the correlation of our dimension with surveyed associations, but is fragile for analysis of historical culture when we can field no surveys.

To identify the connotation of a word on a cultural dimension, we calculate the orthogonal projection of the normalized word vector onto the cultural dimension of interest. Because vectors are normalized, the projection of a word vector onto a "cultural dimension" is equivalent to their cosine similarity. For instance, to determine the gender association for the word "tennis," we project *tennis* onto the gender dimension of (*man – woman*) + (*he – she*) + (*him – her*) +..... In this case, a more negative projection would indicate a more feminine association and more positive values a more masculine association (the signs may be flipped, of course, making positive values reflect feminine associations if masculine terms are subtracted from feminine terms, i.e. by using *woman – man* instead of *man – woman*). By comparing the projections of multiple words on a single cultural dimension, it is possible to compare their connotations within the given cultural category. The process for identifying cultural dimensions and locating the position of words on these dimensions is graphically summarized in Figure 2.[5]

---

[4] Because the cultural category of race is itself multidimensional, its representation in word embeddings is multidimensional as well. While we restrict our analyses to the *black - white* dimension, other word pairs such as *hispanic - white* or *hispanic - black* similarly capture meaningful semantic relations.

____________________________

Figure 2 about here

____________________________

The first panel of Figure 2 shows the construction of a gender dimension by averaging the differences of several gender antonym pairs. The second panel depicts how, by projecting the names of several sports onto the gender dimension, we find that softball and tennis project onto the feminine side of the dimension, soccer is nearly orthogonal to the gender dimension, indicating no strong gender association, and hockey, boxing, and baseball all project masculine. The third panel shows how this process can be repeated for another dimension, in this example class, and how words may be simultaneously positioned along multiple, meaningful cultural dimensions. Furthermore, the angle between these dimensions can be calculated to capture semantic similarity, and it can be evaluated at multiple time points to trace changes in categorical relations. Induced dimensions like gender, class or any other will be approximately orthogonal if they are semantically and contextually unrelated.[6] When the angle between dimensions deviates from 90 degrees, it suggests a meaningful relationship between them, as we demonstrate below.

---

[5] While these low-dimensional cartoons are far different from the actual high dimensional spaces, the distances represented correspond to those in a word embedding learned from a 100 billion words of 21st Century Google News.

[6] We know this from the Gaussian Annulus theorem, that two random points from a $d$-dimensional Gaussian with unit variance in each direction are approximately orthogonal (Blum, Hopcroft, and Kannan 2016).

Our technique of identifying cultural dimensions is closely related to recent work that uses word embedding models to detect bias in texts. Caliskan et al. (2017) demonstrate that a word's position relative to masculine and feminine terms in a word embedding model is strongly associated with that word's associations measured by Implicit Association Tests (IAT) that capture unconscious bias (Greenwald, McGhee, and Schwartz 1998). They use this evidence to argue that word embedding models reveal gendered and racial biases implicit in texts. Bolukbasi et al. (2016) deploy a related approach to neutralize such biased associations in text. Our work builds upon these studies in several ways. First, we show that word position within the embedding model correlate not only with the unconscious associations measured by an IAT, but also with widely shared, consciously revealed associations measured by survey. We argue that the method presented here detects not only hidden biases, but widely shared cultural associations. It is important to note that many associations we find here are indeed biased. For example, "doctor" is consistently found to be more "white" than "black," and "scientist" more "masculine" than "feminine." We argue that word embeddings include bias, however, only because they accurately reflect cultural systems that are themselves rife with bias. Thus, it is rarely in the interest of cultural analyst to "de-bias" a word embedding model as Bolukbasi et al. (2016) propose for job ads and even-handed reportage. Rather, it is by interrogating these biases, as well as the neutral cultural associations present in the models, that the analyst can cultivate an understanding of the multifaceted term meanings and cultural categories deployed in text.

Garg et al. (2017) begin to move in this direction by using word embeddings to study change in gender and ethnic stereotypes over time. By examining change in stereotypes and bias, they recognize them as more than simply distortions of the semantic system, but rather meaningful characteristics that reflect the culture in which texts were produced. Their analysis, however, like Bolukbasi et al. (2016) and Caliskan et al. (2017) remains couched in the analysis of bias rather than cultural categories in general. This paper builds upon and beyond these studies by interpreting the dimensions of embedding models as representative of meaningful cultural categories rather than simply biases, distortions or deficits in the semantic system. We then use these dimensions as tools to illuminate complex cultural relations within a given social context, across contexts, and across time. More broadly, this paper is the first to specifically demonstrate the utility of word embedding models for sociological and cultural inquiry.

Multiple word embedding approaches have become widespread in recent years, but the analyses we present here primarily utilize the CBOW and skip-gram models in *word2vec*, cross-validated with those from *GloVe* (Pennington, Socher, and Manning 2014) and *fastText* (Joulin et al. 2016). The methodological principles outlined here, however, reach beyond neural-network autoencoders and are generally applicable to word embedding models constructed with other algorithms, including Latent Semantic Analysis based on SVD (Dumais 2004) and Bayesian nonparametric approaches (Rudolph et al. 2016).

WORD EMBEDDINGS AND CULTURAL THEORY

Word embeddings are promising tools for cultural analysis because the models they produce align and contend with assumptions in many contemporary sociological theories of culture. One theoretical appeal of word embedding models is the fundamentally relational nature of their meaning representations. Word embeddings are naïve as to what words signify, lacking intrinsic word referents, and position words relative to one another based purely on how they are used in relation to one another. This process of identifying a word's meaning from context resonates with a tradition of practice-oriented theories of language in which word meanings are always understood through usage (Austin 1962; Searle 1969; Wittgenstein 1953). The theory of meaning implicit in word embedding algorithms is well summarized by linguist J.R. Firth's dictum: "you shall know a word by the company it keeps" (Firth 1957).

Word embeddings are also relational in that word meanings are entirely defined by their position relative to other words in the vector space. "Boy" only has meaning in that it is positioned near "man" but closer to "child," near "girl" but closer to "male." At the same time, "male," "child," and "girl" themselves only achieve meaning through their position relative to "boy" and other words in the space. This purely relational approach to modeling meaning parallels a diverse body of cultural theorizing, including structuralist models of meaning developed by Ferdinand de Saussure (de Saussure 1916) and elaborated by Claude Levi-Strauss (Levi-strauss 1963), positing that individual signifiers are arbitrary and acquire meaning only through placement in a complex system of signification. This fundamental insight that meaning is not immanent within words and phrases, but rather coheres within a broader cultural system (Emirbayer 1997) is inherent in any word embedding analysis. Our approach to word embedding analysis further evokes structuralist thinking in the method we present of constructing cultural dimensions using binary antonym pairs. According to structuralist thought, the classification of cultural objects into overarching dimensions such as male/female, clean/unclean, and sacred/profane form the conceptual skeleton upon which culture takes shape (Douglas 1966; Levi-strauss 1963). Our method of constructing dimensions from antonym pairs such as *man − woman* and *rich − poor,* while remaining free of many of the assumptions of structuralism, preserves the basic insight that culture is organized into dimensions defined by binary opposition.

Vector-space representations of culture both resonate and contend with more contemporary relational field theory developed by Pierre Bourdieu. Bourdieusian cultural fields offer a model how individuals, cultural objects, and positions in social structure are located relative one another in structurally homologous "social spaces," with the relations between entities described in terms of "distances" (Bourdieu 1989). Bourdieu frequently represented these social spaces geometrically using the method of correspondence analysis (CA), rendering distances between entities and meaningful dimensions of the field visible by placing them in a two-dimensional plane (Bourdieu 1984). The vector-space models produced by word embeddings similarly position objects relative one another in a shared space based on cultural similarity, but by leveraging the tremendous amount of information contained in a large corpus, word embeddings are able to position words in a semantically-rich, high-dimensional space that need not be reduced to low dimensionality for interpretation. Indeed, the low-dimensional projection of correspondence analysis operationalizes a theory of cultural capital that is itself low-dimensional, that social actors struggle to obtain and maintain dominant positions within a cultural field through a single currency of cultural capital and a single dimension of status-distinguishing tastes and preferences (Bourdieu 1984). By preserving higher dimensionality in a cultural space, word embeddings can facilitate the development and testing of high-dimensional theories of how actors acquire and exploit varied cultural capitals along multiple distinct dimensions of status.

The theoretical implications of moving from low-dimensional to high-dimensional spatial models have been revolutionary in other fields. Sewall Wright, alongside Ronald Fisher and J.B.S. Haldane founded population genetics and constructed the modern synthesis linking genetics to evolution. Wright's low-dimensional cartoon of evolution, however, occurs in two- and three-dimensional "fitness spaces" and posed a fundamental problem for evolutionary theory: how could a population of organisms resting atop a low-dimensional fitness "peak" evolve greater fitness if its progeny needed to dip down into an unfit "valley of death" before climbing to higher global fitness (S. Wright 1932, 1988). Following the discovery of DNA and the acknowledgement that fitness landscapes existed in very high dimensions, the number of possible fitness-neutral evolutionary pathways mathematically exploded and the importance of neutral drift was discovered to play a much more important role than previously conceived (Gavrilets and Gravner 1997). Similarly, the realization that discursive culture exists in high dimensions has theoretical implications for Bourdieusian spaces, which may represent more cultural constraint than actually exists for social actors and entrepreneurs.

The ability of word embedding models to simultaneously locate objects in multiple categories, such as race, class, gender and many others makes them a particularly powerful tool for studying the intersectionality of cultural categories. The fundamental insight of the intersectionality literature is that cultural categories, particularly those of identity, cannot be isolated and understood independently (Crenshaw 1991; McCall 2005). Rather, analysts must always consider ways in which the meaning of cultural categories change as they overlap and intersect one another. Interrogation of the intersection of cultural categories becomes empirically tractable through word embedding models. For example, comparing words that project high on both masculinity and "whiteness" to those that project high on masculinity and "blackness" will reveal how markers of masculinity differ across racial lines within the cultural context in which the texts were produced. This theory that identity is defined by numerous cross-cutting and overlapping categories is itself predicated on a "high-dimensional" model of culture similar to that modeled by Euclidean word embeddings. Indeed, the empirical success of word embedding models to represent cultural dimensions, which we demonstrate below, promotes a radical view of intersectional identity, modeled not as a low-dimensional matrix, but rather a high-dimensional tensor comprised of hundreds or thousands of distinct cultural associations.

Furthermore, word embedding models are well-equipped to discover the heterogeneity of cultural systems by sampling different "voices" in a population. Techniques that model cultural systems with aggregated survey statistics, such as Bourdieu's correspondence analysis, have a tendency to represent cultural associations as homogenous in a population. Bourdieu, for instance, uses correspondence analysis to locate people and cultural objects in a single space reflective of the objective social distances between entities. Nevertheless, Bourdieu recognized that "the vision that every agent has of the space depends on his or his position in that space" and therefore any actual actor's vision will depart from the objective social mapping produced by correspondence analysis (Bourdieu 1989). Word embedding models depict relations as presented in texts produced by a given social group, and hence word embedding models are able capture precisely these visions of social space *from a given perspective*. While word embedding models represent cultural spaces closely akin to those described by Bourdieu, they are spaces as seen from a given vantage point, and therefore have the potential to grant insight into how cultural fields are perceived by distinct social actors and to formally compare these perspectives.

Like Bourdieu's low-dimensional correspondence analysis of survey responses, our use of word embeddings shares much in common with Osgood's semantic differential method, which similarly rates words along cultural dimensions. In the semantic differential method, respondents are asked in an interview to place words on culturally meaningful spectra; for example, is "dictator" closer to "smooth" or "rough?" (Osgood 1952). A key finding from this method is that much of the variance across all dimensions tested can be explained by just three core factors: Evaluation (good vs. bad), Potency (powerful vs. weak), and Activity (lively vs. torpid) (Osgood 1962). Osgood's insight matured within sociology into Heise's affect control theory, which posits that individuals interpret events and plot courses of action by accounting for culturally based affective meanings—or the position of words on stable, culture-specific Evaluation, Potency, Activity (EPA) dimensions (Heise 1979, 1987; Schröder, Hoey, and Rogers 2016). Scholars in this tradition have surveyed individuals in multiple languages and cultures to personally embed thousands of words within culturally specific EPA dimensions.

Despite their similarities, there exist critical differences that distinguish word embeddings from the semantic differential approach. Semantic differential methods have relied upon interview subjects to rate words, but word embedding models utilize text and can sample populations impossible to interview, such as those from the historical past. In addition, each defined dimension must be surveyed individually in the semantic differential approach, whereas word embedding models capture an entire cultural space of associations passively "surveyed" from hundreds of thousands of authors. The semantic differential method presents compelling evidence that much variance in meaning can be explained by a few factors, but the optimization of word embedding models have demonstrated that many more than three dimensions are nevertheless required to model the full variation of words in discourse and effectively perform analogy tasks. Psychologists and computer scientists have recently begun to embed survey responses in high-dimensions for much higher rates of prediction (Jamieson and Nowak 2011; Tamuz et al. 2011; Castro et al. 2009), rather than distilling them like Osgood and Bourdieu into two or three. Moreover, sociologists are often concerned with subtle nuances of meaning and differences along specific dimensions, like gender and race, which may exhibit even more cultural conservation and evolutionary significance than EPA dimensions. In summary, EPA may be useful to coarsely bin concepts, but the higher dimensionality of word embedding models enable fine-grained distinctions, associations, distances and the discovery of emergent cultural dimensions.

DATA & METHODS

We draw upon multiple, varied data sources to demonstrate the power of word embeddings for cultural analysis. First, to establish the ecological validity of results produced by word embeddings we conducted a survey that measures associations between everyday objects and the cultural categories of race, class, and gender. We compare results from this survey with associations calculated using word embedding models trained on contemporary sources of text. These contemporary sources include three widely-used, publicly-available pretrained models and a fourth we trained on the publicly available Google Ngrams corpus. Following this validation, we use Google Ngrams to conduct cross-national and historical analyses.

*Survey of Cultural Associations*

In order to establish a basis of comparison between individuals' actual reported cultural associations and the associations as represented in the word embedding model, we fielded a survey of cultural associations to a total of 398 respondents on Amazon Mechanical Turk. The survey was fielded in two waves, the first in 2016 and the second in 2017, and was open only to Mechanical Turk users located in the United States. Although our sample cannot be said to be representative of the general population of the United States, responses to basic demographic questions indicate wide diversity in age, gender, and racial composition in the sample. To improve representativeness, we apply post-stratification weights to the sample, weighting on race (white, black, or other), education (bachelor's degree or less), and sex (male or female). The results presented here include post-stratification weighting, but supplemental analyses available upon request show that unweighted models produce substantively similar findings. This survey and the weighting procedures are described in greater depth in Appendix A.

In the survey, respondents were asked to rate a number of items on scales representing association along gender, race, and class lines. All questions followed the format, "On a scale from 0 to 100, with 0 representing *very feminine* and 100 representing *very masculine*, how would you rate a *steak*?" For measuring race and class associations, the survey posed similarly worded questions, replacing "feminine" and "masculine" with "White" and "African American," or "working class" and "upper class" respectively. Respondents were asked to place 59 different items on each of the three dimensions of race, class, and gender. A full list of items asked on the survey is available in the appendix (Table A1). Words were selected in seven topical domains: occupations, foods, clothing, vehicles, music genres, sports, and first names. A diverse array of topical domains were chosen to test the capacity of word embedding models to detect cultural associations across very different subjects. Specific terms were selected within each topical domain to ensure high variance across dimensions[7]. We calculate the weighted mean of responses for each item and use these means as our estimates of a general cultural association. The end product is thus a rating between 0 and 100 on a gender dimension, a class dimension, and a race dimension for each of the 59 words listed in Table A1.

*Word embedding data*

To represent contemporary cultural associations, we utilize three widely-used, publicly-available word embeddings and a fourth word embedding model that we construct using Google Ngram text. The first model we use to represent contemporary cultural associations is a word embedding trained using the *word2vec* algorithm on 100 billion words scraped from Google News articles published by American news outlets (Mikolov et al. 2013). This embedding is well-suited for representing widespread contemporary cultural associations for three reasons. First, news media covers a great breadth of topics and is intended for a wide audience, and therefore must speak with a voice interpretable to the general public. Second, print and online news are narrative genres, repeating stories that contain the kind of implicit associations likely to reflect common experiences within the culture. Third, this embedding is trained on a particularly large corpus, and previous tests with word embeddings have demonstrated that meaning accuracy, measured as the success rate at solving analogies, greatly improves as the corpus size increases (Mikolov et al. 2013). The second model, also widely used and publicly available, was produced by the *GloVe* algorithm, which accounts for both local and global dependencies, and is trained on a corpus collected as part of the Common Crawl, a broad scraping of millions of webpages (Pennington, Socher, and Manning 2014). It similarly has the benefits of size and breadth of subject matter, and is here used to test the robustness of word embeddings for capturing cultural associations with a related word embedding algorithm on a different corpus. The third model is a publicly available embedding trained using the *fastText* word embedding algorithm on text from Wikipedia (Bojanowski et al. 2016). The *fastText* embedding algorithm is a modification of *word2vec* that accounts for sub-word information such as prefixes and suffixes. The fourth embedding we validate uses *word2vec* on Google Ngram text from 2000-2012. The Google Ngram corpus is the product a massive project in text digitization, in collaboration with thousands of the world's libraries, which distills text from 6% of all books ever published (Lin et al. 2012; Michel et al. 2011). Any sequence of five words that occurs more than 40 times over the entirety of the scanned texts appears in the collection of 5-grams, along with the number of times it occurred each year. Because word embeddings require local context to determine the meaning of words, we limit our analysis to the collection of 5-grams, and exclude data on the occurrence of 4-grams, 3-grams, 2-grams, or single words. All characters were converted to lowercase in preprocessing in order to increase the frequency of rare words. Although the Google Ngrams corpus does not represent one single, identifiable genre or voice, it spans such a wide variety of genres including novels, government documents, academic texts and technical reports, that this sheer breadth enables it to faithfully represent the most pervasive and widely shared cultural markers of the time period.

---

[7] First names were sampled from lists of names found to be most predictive of belonging to an African American and most predictive of belonging to a non-Hispanic white for each sex from data of all children born in California over the period 1961–2000 (Fryer and Levitt 2004; Levitt and Dubner 2005). Terms in other domains were selected based on known race, class, and gender markers that have been examined in previous literature.

  After validating our method on contemporary text, we again draw upon Google Ngrams for our analyses of historical cultural change and cross-cultural comparison. To investigate how the categories of gender and class transformed in American culture over the 20th century, we utilize 5-grams of text from books published in the United States between 1900 and 1999. We divide the corpus by decade, training separate models on texts from 1900-1909, 1910-1919, and so on through 1990-1999, resulting in ten independently constructed word embedding models. By comparing these models side-by-side, we are able to trace out patterns of macro-cultural change over this 100 year period.

To demonstrate the utility of word embedding models for cross-cultural comparison, we train models on texts published at the turn of the 20th century in the United States and in the United Kingdom. The U.S. and U.K. serve as particularly useful comparison cases because they share a common language yet exhibit distinctive cultural differences. We chose to analyze text from the turn of the 20th century for two reasons. First, one advantage of word embedding models is that they are able to construct detailed models of cultural systems that have left a textual trace but are no longer available for direct observation. Thus, word embeddings are uniquely well-equipped for comparing cultural groups from the past. Secondly, processes of globalization and institutions of meritocracy and status mobility greatly accelerated during the 20th century, particularly after the first World War. By comparing the United States and the United Kingdom prior to World War I, we are able to gain insight regarding how these countries' cultures historically differed prior to this explosion of international exchange and status shuffling. The text we use for this cross-cultural comparison is also drawn from the Google 5-gram corpus. Google Ngrams preserves metadata on books' country of publication, making it possible to separately train models on texts from the U.S. and U.K. We therefore train two sets of models, one on text published in the United States and another on text published in the United

Kingdom, both between 1890-1910. We use a twenty-year window to ensure enough text to capture subtle but meaningful cultural traces. We then compare those models to gain insight into cross-national cultural differences in gender and class association from the turn-of-the-century period.

Although the Google Ngram corpus has been subject to criticism because the composition of the corpus in a given year may not be representative of total literary output (Pechenick, Danforth, and Dodds 2015), we present evidence suggesting that this issue is less problematic for word embedding approaches than it may be for analyses based on raw word frequency counts. Because the positioning of words in a word embedding model is predicated on consistently shared context with other words rather than frequency of occurrence, it makes little difference if a word is talked about more or less – its position will only change if it is talked about *differently*. Our analyses of contemporary text show that the gender, class, and racial associations for a wide range of objects are similarly captured whether the word embedding is trained on books, news, Wikipedia entries, or webpages. This suggests that the gender, class, and racial associations of everyday items with which we are concerned in this paper are diffuse and sufficiently shared to exhibit a similar character across textual genres and likely would not vary markedly with shifts in the composition of Google Ngrams over time. Furthermore, in Appendix B we construct empirical tests that demonstrate the accuracy of the gender dimension for correctly classifying the gender of popular first names across the 20th century, and find that accuracy remains high across all ten decades.

At multiple points in our analysis, we seek to evaluate cultural dimensions not only with respect to each other, but also the full range of other cultural dimensions. Neural network autoencoders are "unsupervised" machine learning models in that they are not trained to match the associations of human experts (Evans and Aceves 2016). As detailed above, unsupervised approaches to text analysis also include topic models and semantic networks, which automatically produce complex representations of word associations over all available data. Nevertheless, analysts still typically sample interpretations from these complex artifacts, often by selectively noting salient topics, word linkages or word vector proximities. This has led to reasonable allegations that researchers conveniently cherry-pick explanations to confirm pre-existing opinions (Nickerson 1998). To avoid this temptation, we attempt to systematically sample from the space of possible interpretations by extracting a list of 1775 antonym pairs from WordNet, an electronic lexical database containing of a thorough catalogue of English words, their meanings, and relations to one another (Miller 1995), including antonymy. We then compare focal dimensions with cultural dimensions inscribed by a nearly exhaustive list English antonyms, and restrict ourselves to dimensions posting the smallest angle with respect to those focal. While this approach necessarily requires a commitment to a particular conceptual dimension such as gender, it enables us to systematically "interpret" that dimension with *all* others.

*Statistical Significance of Distances and Associations*

We propose well-established bootstrapping and subsampling methods to nonparametrically demonstrate the stability and significance of word associations within our embedding model. These allow us to establish conservative confidence or credible intervals for both (a) distances between words in a model and (b) projections of words onto an induced dimension (e.g., *man-woman*). If we assume that the texts (e.g., newspapers, books) underlying our word embedding model are observations drawn from an independent and identically distributed (i.i.d.) population of cultural observations, then bootstrapping allows us to estimate the variance of word distances and projections by measuring those properties through sampling the empirical distribution of texts with replacement (Efron and Tibshirani 1994; Efron 2003).[8]

If the corpus is very large, however, then we may take a subsampling approach, which randomly partitions the corpus into non-overlapping samples, then estimates the word-embedding models on these subsets and calculates confidence or credible intervals as a function of the empirical distribution of distance or projection statistics and number of texts in the subsample (D. N. Politis, Romano, and Wolf 1997). Subsampling relies on the same i.i.d. assumption as the bootstrap (Dimitris N. Politis and Romano 1992, 1994). For 90% confidence intervals, we randomly partition the corpus into 20 subcorpora, then calculate the error of our embedding distance or projection statistic $s$ for each subsample $k$ as $B^k = \sqrt{\tau_k}(s^k - \bar{s})$, where $\tau_k$ is the number of texts in subsample $k$, $s^k$ is the embedding distance or projection for the $k_{th}$ sample, and $\bar{s}$ is the full sample estimate. The 90% confidence interval spans the 5th to 95th percentile variances, inscribed by $\bar{s} - \frac{B^k_{(19)}}{\sqrt{\tau}}$ and $\bar{s} - \frac{B^k_{(2)}}{\sqrt{\tau}}$ where $\tau$ is the number of texts in the total corpus. As with bootstrapping, a 95% confidence interval would require 40 subsamples; a 99% confidence would require 200 (.5[th] to 99.5[th] percentiles). A great benefit of bootstrapped and subsampled confidence intervals is that they reflect how robust an association is across texts. If a word occurs only rarely or is used in a diffuse set of very distinct contexts, the word's position in the vector space will be radically different between subsamples and therefore will produce larger confidence or credible intervals. On the other hand, words that are frequently used in very consistent contexts will hold more stable positions across the subsamples and hence produce smaller confidence or credible intervals.

---

[8] Operationally, if we wanted to bootstrap a 90% confidence interval of a word-word distance or word-dimension projection, we would sample a corpus the same size as the original corpus, but with replacement, 20 times, and estimate word embedding models on each sample. Then we take the 2[nd] order (2[nd] smallest) statistic $s_{(2)}$—either distance or projection—as our confidence interval's lower bound, and 19[th] order statistic $s_{(19)}$ as its upper bound. The distance between $s_{(2)}$ and $s_{(19)}$ across 20 bootstrap samples span the 5[th] to the 95[th] percentiles of the statistic's variance, bounding the 90[th] confidence interval. A 95% confidence interval would span $s_{(2)}$ and $s_{(39)}$ in word embedding distances or projections estimated on 40 bootstrap samples of a corpus, tracing the 2.5[th] to 97.5[th] percentiles.

RESULTS

To determine the relative importance of cultural dimensions in the embedding space, we calculate the total amount of variance explained by the gender, race and class dimensions across all words contained in the space. Using the Google News embedding model, we find that gender, class and race explain a small percentage of the total variance in the space: gender explains 0.57%, race 0.48%, and class 0.47%. These values are modest not because gender, race, and class do a particularly poor job of capturing semantic difference, but because words have so many valences that render across a 300 dimensional embedding space, no single dimension can capture a large fraction of the variance. Using principal components analysis, we find that the dimension capturing the most possible variance in the space explains only 3.65% of its variance. These findings suggest that the rich semantic spaces produced by word embeddings cannot be reduced to very low dimensionality—such as Osgood's three-dimensional EPA model—without losing a majority of cultural information.

*Survey validation of models*

We begin our analysis by constructing the cultural dimensions for gender, class, and race, which will be used in all subsequent analyses. Table 1 lists the pairs of words we use to approximate gender, class, and race dimensions. As described above, we select word pairs such that the difference between each word in a pair is a "step" along the dimension of interest. For example, the difference between *richest* and *poorest* is a step along the class dimension, and should be a similar in direction to the difference between *affluence* and *poverty*. The lists presented in Table 1 are not exhaustive; other pairs of words could be added or substituted for the pairs we have included. Averaging between six and ten pairs of words for each dimension, as we have done here, produces a closer approximation to the cultural dimension of interest than any one word-pair, as they post higher correlations with the survey described below. Cultural dimensions are calculated by simply taking the mean of all word pair differences that approximate a given dimension, $\frac{\sum_{p}^{|P|} \vec{p_1} - \vec{p_2}}{|P|}$, where *p* are all word pairs in relevant set *P*, $\vec{p_1}$ and $\vec{p_2}$ are the first and second word vectors in each pair.

---

Table 1 about here

---

Having recovered dimensions of the word embedding model that correspond to cultural categories of gender, class, and race, we proceed by validating these associations by comparing them with associations measured in the survey. We test the correspondence between results from the survey and the word embedding model by calculating the Pearson's correlation between a word's mean rating on a given scale in the survey and the projection of the word vector on the respective cultural dimension in the embedding model. The correlation is calculated using the 59 words that are listed in Table A1. Results are presented in Table 2.

______________________________

Table 2 about here

______________________________

The first column of Table 2 displays the correlation between a word's average rating on the gender scale from the survey and the word's projection on the gender dimension for four embedding models. In the first row of Table 2, we see that the correlation between a word's gender rating on the survey and that same word's projection onto the gender dimension in the Google News word embedding is 0.88. This correlation is impressively high and attests to how well a gender dimension elicited from the word embedding model corresponds to contemporary individuals' understandings of masculinity and femininity. We find similarly high correlations between the projections on the gender dimension of the *GloVe* embedding trained on the Common Crawl text, the *fastText* embedding trained on Wikipedia text, and the *word2vec* embedding and Google N-grams, at 0.90, 0.87 and 0.78 respectively. These strong correlations suggest that projections on the gender dimension of a word embedding model can serve as a proxy for the gender associations that would be reported in a survey by respondents sharing a similar set of cultural understandings to those presented in the text.

The second column of Table 2 presents correlations between survey response and word vector projection for class. We see that associations are generally weaker than they are for gender, but remain substantial, ranging between 0.40 and 0.60. Four possible explanations for the weaker association are (1) greater disagreement among those surveyed regarding race and class associations (average standard deviation=23.1) than gender (SD=20.8); (2) greater disagreement among authors of texts regarding class associations than gender; (3) a disjunct between class associations present in the texts and those understood by survey respondents; and (4) measurement error, both in the embedding and the survey – e.g., the survey may not elicit associations of class as sharply as gender, and *rich – poor* or *affluent – impoverished* may not approximate the class dimension from texts as perfectly as the *man – woman* or *he – she* captures the gender dimension.

The third column shows correlations between the outputs of word embedding models and the survey for racial associations. We see that the Google News embedding does excellently, with a correlation of 0.70, and the Common Crawl and Wikipedia embeddings do moderately well with correlation of 0.42 and 0.49 respectively, while the Google N-grams corpus does relatively poorly, with a correlation of only 0.17. The failure of the n-grams model to pick up racial associations may be a product of the texts included in the corpus. The subject matter of news articles and general internet postings may be imbued with more racial association than the Ngrams corpus, which contains significant non-fiction, including technical reports and scientific publications without narrative content that would invoke ambient, contemporary racial associations within that embedding model's projections. Additionally, as noted at the bottom of Table 1, the Google Ngrams text were entirely reduced to lowercase in preprocessing, which decreased the available number of antonym word pairs for constructing the race dimension from six to four, possibly resulting in decreased accuracy of the dimension.

In Table 3 we provide a more detailed summary of the correspondence between associations produced in the word embedding model and those reported by survey respondents. For all pairs of words that have a statistically significant difference in mean survey rating (p<0.01) for gender, class, and race associations within a substantive domain, we calculate the proportion of those pairs that are correctly ordered by the word embedding model trained on Google News text. For instance, if "steak" is significantly more masculine than "salad" in the survey, we test if *steak* more masculine than *salad* in the embedding, and then calculate the percentage of all such pairs of words that are correctly matched.

______________________________

Table 3 about here

______________________________

We see in Table 3 that within most substantive domains, the rate of correct classification is above 80% and in many cases above 90%. It is also clear that embedding does a better job in domains with stronger cultural associations. For instance, there is very little difference in racial association between the clothing items included in the survey (standard deviation of 4.68) and in this domain the embedding has a low 51.8% rate of matching the survey. In first names, however, where signals are stronger (standard deviation of 32.46) the same dimension of the word embedding correctly matches 100% of differences in the survey. It is also important to note that we should not expect a 100% correspondence in all fields because the Google News corpus does not perfectly correspond to the cultural associations shared in the population sampled in this survey; presumably the authors of news articles are not culturally identical to a sample of Mechanical Turk respondents. Therefore, we would expect a word embedding trained on a corpus that matched the cultural associations of the survey respondents to exhibit a much higher correlation with survey results than we observe here.

A visual depiction of the correspondence between word embedding models and our survey of cultural associations is presented in Figure 3. Figure 3 reveals how several music genres – jazz, rap, opera, punk, techno, hip hop, and bluegrass – are arrayed on the cultural dimensions of race and class by survey and word embedding, with the average survey rating of a word depicted in black and projection in gray. Comparing survey output to word embedding projections, we see striking similarity in the relative positions of words. By both methods, opera holds the association of being at once both high class and white. Techno, punk, and bluegrass are similarly white, but of distinctly lower class than opera. On the right end of the panel, we see that the music genre associated with both African Americans and high class is jazz, whereas hip hop and rap tend working class. Projecting words simultaneously into multiple dimensions, it is clear how word embeddings can be used to examine intersectionality by revealing how class markers vary across racial lines.

_______________________

Figure 3 about here

_______________________

It is important to note that word embedding models capture cultural associations present in text, but this does not always reflect material or historical realities. In some cases embedding models succeed in capturing historic realities; hip hop, rap, and jazz are truly historic products of African American culture and they carry a clear African American association in the word embedding. Music historians, however, also trace the origins of techno to African American artists based in Detroit in the 1980s (Reynolds 2013), yet the embedding model locates techno's contemporary association at a relatively white position on the racial dimension. Comparing this to survey results, we see that the survey responses similarly fail to reflect the historic association of techno music with African Americans. Thus, both survey results and embedding models suggest that in spite of African Americans' historic involvement in the invention of techno music, techno currently carries little sign of African American association among the broad public. This case serves as a useful demonstration of how embedding models are able to successfully capture cultural associations, but how these cultural associations may themselves be limited, distorted or biased representations.

Here we focus primarily on race, class, and gender, but any number of meaningful cultural categories can be interrogated using the method we outline for recovering the cultural dimensions of a word embedding model. For example, creating a liberal/conservative political ideology dimension is straightforward and can be quickly accomplished without collection of additional data, simply by averaging antonym pairs that trace this political dimension. Figure 4 maps a number of objects of lifestyle politics in a two-dimensional space, with the x-axis representing ideological association and the y-axis representing class association, calculated using the *word2vec* Google News embedding[9]. The arrangement of words on these dimensions depicts not only how ordinary items, places, and social groups are charged with political connotation, but also how political markers differ between socio-economic levels. In the top left of Figure 4, we see markers of upper-class liberalism include "Subaru," "Prius," "New_York," and "latte," whereas the lower left corner contains "hippie," "feminist," and "union" as markers of the Left lacking association with affluence. In the top right we see markers of high-class conservatism: "golf," "steak," and "business," while less affluent markers of conservatism comprise "Dodge," "evangelical," "rancher," "pickup," and "Texas." These associations all closely mirror those described in popular media and scholarly work on lifestyle politics, graphically depicting not only how everyday items have ideological connotation, but also how markers of ideology are stratified by class (DellaPosta, Shi, and Macy 2015). To provide additional evidence of word embeddings' ability to capture political associations, we show in Appendix B1 that the Google News embedding is able to categorize a set of recent high-profile politicians as Democrats of Republicans with near perfect accuracy.

---

[9] In Figure 5, the political political dimension is an average of the following five word differences: *liberal-conservative, liberals-conservatives, Democrat-Republican, Democrats-Republicans, Democratic-Republican*. The class dimension is the same as in previous analyses.

———————————

Figure 4 about here

———————————

*Analyzing Historical Change in Cultural Dimensions*

Having established a strong correspondence between cultural dimensions identifiable within a word embedding model and those present in the contemporary collective consciousness, we proceed to analyze word embedding models trained on text produced in previous decades to reconstruct and examine past systems of cultural association. For this task, we once again draw upon text from the Google Ngrams corpus, limiting our analysis to texts published in the United States between 1900 and 1999, dividing the corpus by decade and training a model on each 10-year period independently with the *word2vec* algorithm. By training separate word embeddings on text from each decade of the 20$^{th}$ century, then projecting words along gender and class dimensions in each embedding, we trace out broad patterns of cultural change and reproduction regarding gender and class associations over the past century. Investigating race over time, while possible with word embeddings, would present unique challenges. Because words used to denote racial groups shifted repeatedly over the 20th century, words used to construct the race dimension in the embedding would also need to change from decade to decade. Undertaking this task effectively would require a deeper engagement with linguistic shifts in racial terminology than is feasible within the space constraints of this paper. Furthermore, findings from Table 2 suggest that the Google Ngrams corpus successfully captures gender and class associations but does relatively poorly in recovering racial associations. We thus limit our historical analyses to an examination of gender and class.

    We illustrate the technique of tracing historic evolution in word meaning by investigating change over the 20th century in the class and gender associations of three occupations that exhibit distinctly different trajectories: nurse, journalist, and engineer. Results are presented in Figure 5. We note that most associations remain stable from one decade to the next; "nurse" is consistently feminine, "engineer" remains masculine, while "journalist" changes considerably but gradually over several decades. The general stability from decade to decade enables us to compare these models trained on texts from sequential periods. Nevertheless, the specific characteristics of each occupation's trajectory is informative and unique. "Nurse" shows a consistent, incremental movement from strong to weaker feminine association. Conversely, "engineer" traces a slow movement towards decreasing masculinity. "Journalist," changes gender connotation more dramatically, beginning as a masculine occupation in the beginning of the twentieth century, but shifting to an increasingly feminine career by century's end. The changes in gendered association of "engineer" and "nurse" are consistent with a great deal of evidence documenting decreasing occupational gender segregation across the 20th century (Blau, Brummund, and Liu 2013; Reskin 1993) and in fields of engineering and nursing in particular (D'Antonio and Whelan 2009);

National Science Foundation 2016). The shift from masculine to feminine association for journalism corresponds to a real increase in women's representation in journalism, but labor statistics indicate that journalism remained male dominated 20$^{th}$ century's end (ASNE 2017). Gender scholars have noted a feminization of the culture of journalism in late 20th century, however, as the media outlets began to cater to female consumers (Oliver 2014; Van Zoonen 1998). This cultural association is corroborated by survey results, in which "journalist" was on average rated slightly feminine. Once again, we caution that word embeddings models, by distilling information from local word co-occurrences, do not provide direct information about material, on-the-ground realities. Rather, they capture semantic associations that permeate speech and text, which remain fundamentally cultural.

———————————

Figure 5 about here

———————————

The second panel of Figure 5 traces the evolution of class association for the same three occupations. It is immediately clear that class shifts are generally distinct from changes in gender association. As nursing became less feminine, it also rose in class. "Journalist" became dramatically more upper-class while at once becoming more feminine. This swift movement from middle to upper-class over in the mid-twentieth century is in line with historical accounts of the professionalization of journalism, a process in which journalists ceased to be scrappy, uneducated reporters from middle- and working-class backgrounds, but increasingly affluent, well-connected writers educated at elite colleges (Hess 1981). As journalists transitioned from what Max Weber described in 1918 as "a sort of pariah class" (Weber 1948) to a position of status and prestige by the 1980s, we see the class association of journalist in the word embedding models rise accordingly. "Engineer" is consistently positioned in the middle of the spectrum between upper-class and working-class. Unlike professionals in medicine or law, U.S. engineers were from historically lower class backgrounds and failed to organize and exclude entry through requiring a liberal education as prerequisite—most engineers only obtain a single degree (Collins 1979). Moreover, despite being well paid, engineers' work in mechanical and technical operations often puts them in closer association to blue-collar work than white-collar professionals such as journalists.

Word embedding models not only have the capacity to trace the positions of words along cultural dimensions over time, but also make it possible to holistically evaluate the relationship between cultural dimensions by calculating the angle between them. Dimensions with similar meanings, such as *strong-weak* and *big-small* will tend to have high cosine similarity, meaning they approach parallel and exhibit high correlation between the projection of individual words on each dimension. Conversely, unrelated dimensions, such as *strong-weak* and *hot-cold*, are orthogonal and word projections do not covary. By evaluating the changing angle between cultural dimensions over time we track historical shifts between categories. To demonstrate this approach, we trace the relationship between class and gender over the 20$^{th}$ century by calculating the cosine similarity of the angle—and its arccosine, the angle's degree—between those dimensions. We also calculate the Pearson correlation between the gender projection and class projection for all words in the embedding in each decade to display how non-orthogonal cultural dimensions translate into a correlation between words' positions on the two dimensions.

The angle between gender and class dimensions for each decade of the 20th century is presented in Figure 6 and traces an intriguing pattern; class and gender hold a strong association in the U.S. at dawn of the 20th century, with femininity associated with higher class and masculinity lower class. This association remains stable through the first half of the 20th century, but quickly diminishes to unrelated orthogonality in the final decades. This trend is at first glance surprising. Given that women entered the workforce at growing rates in the second half of the century, it would be reasonable to hypothesize that femininity would take on a higher-class association at this time. Yet the results presented in Figure 6 suggest exactly the opposite, that femininity was more synonymous with high class *before* women entered the workplace.

_______________________

Figure 6 about here

_______________________

To shed more light on this trend, we follow the relationship between class and other key cultural dimensions over the 20th century, illuminating a broad pattern of defeminizing class associations.[10] In Figure 7 we see that in the beginning of the 20th century, high class was associated with the feminine qualities "beautiful," "graceful," and "fine," whereas lower-class was marked with masculine qualities "ugly," "awkward," and "coarse." However, over the course of the century, the class became increasingly orthogonal to the *beautiful-ugly*, *graceful-awkward* and *fine-coarse* dimensions. In their place, class came to be more closely synonymous with the categories of *employed-unemployed* and *educated-uneducated*, with the upper class being increasingly identified as educated and employed and the lower class as uneducated and unemployed. In short, these changes imply that over the course of the twentieth century, class came to have less to do with beauty and grace and more to do with education and employment. This macro-historical trajectory can be understood within the broader historical narrative of the decline of the aristocratic archetype of wealth and its replacement with an image of the high class as educated technocrats at the helm of commerce and industry.

---

[10] We construct cultural dimensions for each of the antonym pairs by following the same process described in the previous section - taking the difference between the two word vectors and normalizing the resulting vector.

_____________________

Figure 7 about here

_____________________

*Cross-National Comparison: United States and Great Britain*

Next we demonstrate how word embedding models can be used as a powerful tool for cross-cultural comparison. Just as historic cultural changes can be investigated by comparing embeddings trained on texts produced in multiple time periods, cross-cultural differences can be analyzed by comparing embeddings trained on texts generated by different social groups. In the following analyses, we conduct a comparison of gender and class associations between the United States and the United Kingdom during the pre-war period of 1890-1910, drawing once again from Google 5-grams text in order to get a broad sampling of discourse that represent diffuse and widespread cultural associations shared by broad literary publics in each country.

We begin our cultural comparison by examining how other cultural categories relate to gender and class in the United States and Great Britain in this time period. Namely, we discover for each country the dimensions most synonymous with gender and with class, and compare these results to find differences cross-national differences in how these categories are conceptualized. From the complete list of WordNet antonyms, we limit our analysis to categories more frequently invoked, listing pairs of antonyms in which both words are within the 20,000 most frequently occurring words in the given country's texts. This subset produces lists of 442 pairs for the United States and 428 pairs for Great Britain. We construct a cultural dimension corresponding to each antonym pair and calculate the cosine similarity between each of these cultural dimensions and the class and gender dimensions.

_____________________

Table 4 about here

_____________________

We present a list of the five cultural dimensions closest to gender for the United States and the United Kingdom in Table 4, which suggest striking similarities between the two cultures at the turn of the 20$^{th}$ century. Of the five dimensions closest to gender in U.S. text, four are also among the top five for Great Britain. Masculinity in both contexts associated with being more rugged, loud, tough, and bold, while femininity was associated with delicateness, softness, tenderness and timidity. Comparing dimensions closest to class, we once again see strong similarities between the two English speaking nations: *weak-strong* is closest to class in the U.S., and the third closest in Britain, while *fortunate-unfortunate* is second closest to class in the U.S. and fourth closest in Britain. Subtle distinctions in the class dimension, however, reveal important differences between the two cultures. The dimension most highly correlated with class Great Britain is *bare-covered* at cosine similarity of -0.255 (104.8º), whereas cosine similarity between bare-covered and class in the U.S. is much closer to orthogonal at only -0.143 (98.2º). We also see that employed-unemployed is highly correlated with class in Britain, but that category would only come to be highly correlated with class in U.S. in the second half of the twentieth century, following different timelines of industrialization for the United States and the Great Britain. By 1900, 39% of the American workforce worked in agriculture (United States. Bureau of the Census 1975), compared with only 7% in Britain (Mitchell 1971).

While the United States and the United Kingdom shared similar basic conceptions of gender and class at turn of the 20th century, the specific markers that carried gender and class connotations differed markedly between the countries. We examine differences in class and gender associations between the U.S. and U.K. by comparing word projections on these dimensions in the two embeddings. In the top panel of Figure 8, we display a comparison of word locations on the class dimension for the United States and Great Britain in 1890-1910, revealing important differences in the class connotation of these items. The top-left quadrant of the figure shows that "rank" and "nobility" carry stronger upper-class association in Britain than the U.S. We also see that "colonies" has a moderately strong upper-class association in Britain, which is not present in the U.S. While British colonies themselves were frequently impoverished by the process of colonial extraction, the colonies still represented a source of riches and wealth for the British Empire. The United States, on the other hand, had much less economic stake in colonies. American associations with colonies are more likely to be in reference to the United States' own colonial past than to the economic fruits of foreign colonies. The top-right quadrant displays words of high class connotation in the United States. We see that "cotton" has a strong connotation of affluence in the U.S. not present in the Great Britain embedding. This is reasonable given that agricultural production, particularly of cotton, had for decades been the backbone of the American South's economy. The word "wilderness" has an upper-class association in U.S.' text whereas it has a moderately lower-class connotation in Britain. This difference reflects the unique position of wilderness in the American imagination. Historians have recognized an identification of the American wilderness as a bountiful frontier to be settled for agriculture or harvested for timber. With the founding of the first National Parks at the end of the 19th century, wilderness was becoming widely perceived as a national treasure in itself (Nash 2014).

______________________

Figure 8 about here

______________________

The bottom-left quadrant of Figure 8 contains words with lower-class connotation in Great Britain. We see that "commoner" is a strong marker of lower-class in Britain, while more neutral in America. Taken alongside the particularly high-class ratings for "nobility" and "rank" in Britain, this firm lower-class connotation for "commoner" suggests a stronger correspondence between social rank and class in Britain than the United States, where economic success was imagined available to members of any social stratum as suggested by widespread popularity of rags-to-riches novels and the myth of the American Dream (Peña 2016; Weiss 1969). This egalitarian framework is also evident in class associations of "worker" and "farmer," both marked as lower class in Britain but not America. Nevertheless, the United States exhibits a profoundly lower-class connotation along racial lines, here captured by the word "negro" (the most prevalent term of the period for people of African descent in both the U.S. and U.K.), which suggest that at the turn of the 20th century, U.S. conceptions of blackness and poverty were intertwined in a uniquely powerful way. Thus, while the American texts suggest a cultural egalitarianism in class associations for "worker," "farmer," and "commoner," our findings imply that race defied this egalitarian tendency, remaining a sharper and more definitive marker of lower class in the United States relative to Great Britain. One final difference of note is the class connotation of the word "tenement." Although overcrowded urban housing is in no way particular American cities, the tenement was a uniquely American structure at the turn of the 20[th] century, and the term was not used to refer to low-income urban housing in Britain (Wright 1983). Matching this history, we see that "tenement" is sharply marked as lower class in the United States, yet carries no such association in Great Britain.

Our comparison of the closest dimensions to gender in the U.S. and Britain, 1890-1910, suggested that the overall category of gender appears quite similar in the two countries, but difference in masculine/feminine connotations for individual items suggest how its application varied. To demonstrate the capacity of word embedding models to pick up subtle cultural associations, we show evidence from a substantive area where American and British cultural histories diverge but where the gender associations are likely to remain implicit: the understanding of foreign, colonized territories. Longstanding theories of culture and colonialism have pointed to a feminization of colonized people and places (Said 1978). We bring new evidence to bare on this widely theorized association by locating the names of continents and countries subject to British colonization on a gender dimension in British and American word embeddings. Results are presented in the second panel of Figure 8, with higher points on the Figure representing more masculine associations. We see that both British and American authors self-portray as highly masculine; in Britain text "england" is the most masculine term in the selection and in American text "u.s.a." is most masculine. Next we see that in the U.S. and Great

Britain, associations of other Western powers are also relatively masculine; "England" is masculine in the U.S., "U.S.A." is masculine in the Britain, and "Europe" is masculine to both. For "Africa," "Asia," and "India," however, sites of intensive British colonization, we see strong feminine associations in British text more neutral gender associations in America. China, which was not colonized but was subject to coercive trade policies by both the United States and Great Britain in the 19th century (Cohen 2010), posts a strong feminine association in both countries' texts. These results corroborate the thesis that colonized territories of the East were culturally depicted and understood as feminine by the West. The findings we present here further suggest that feminization of these key sites was particularly pronounced in Great Britain, the colonizing nation. The United States, which historically had had a lesser role in global colonization, does not share the strong feminine connotations of Eastern locales such as Asia, Africa, or India, with which it did not exert $19^{th}$ century dominance.

DISCUSSION:

In this paper we present modern word embedding models, particularly those produced using the algorithm *word2vec*, as a productive method for the analysis of cultural categories and associations. By representing the relationship between words as the relationship between vectors in a high-dimensional vector space, word embedding models distill vast collections of text into a singular representation while still preserving much of the richness and complexity of the semantic relations for systematic interpretation. We describe how dimensions of the word embedding models correspond closely to "cultural dimensions" such as race, class, gender or liberalism/conservatism, and how examining the position of words arrayed upon the salient cultural dimensions of a word embedding can illuminate patterns of association and classification in a given cultural system. Furthermore, we show that by calculating the angles between the cultural dimensions themselves, analysts can gain insight into the culture-wide relations between categories.

We provided empirical demonstrations of how word embedding models can be utilized to interrogate the relations between words within a given cultural system, but also to draw close comparisons between multiple cultures. In this paper, we presented analyses comparing cultural relations surrounding gender and class across distinct decades of the 20$^{th}$ century, tracing slow but persistent shifts in the relation between gender and class in American culture. We also demonstrate the utility of word embedding models for cross-cultural analysis by comparing embeddings trained on texts produced in the United States and Great Britain at the turn of the 20$^{th}$ century, revealing subtle differences in meaning for a number of cultural markers.

The full range of potential applications for word embedding models for the analysis of culture reaches far beyond the examples presented in this paper. Following the general approach piloted here, analysts could use word embedding models to compare the cultural systems represented between literary genres, texts produced by different authors, or texts written in different languages (Lev, Klein, and Wolf 2015). A wide array of social collectives, including scientific disciplines, political elites, and contributors to online forums, can be analyzed and compared by training word embedding models on the text they produce. Furthermore, while this paper has focused on insights produced by identifying, extracting or comparing "cultural dimensions" from the vector space, we do not maintain that this is the only method for utilizing word embedding models to advance social science. Higher than one-dimensional comparisons could be made by plotting the words within a space inscribed by multiple "archetypes" (Lim and Harrell 2015). Simply calculating the proximity of word vectors can also provide a strong indicator of the similarity or distance between word meanings (Kulkarni et al. 2015). Word embedding models can further be used to classify and predict which group produced a text, given multiple corpora produced by distinct social groups (Taddy 2015). Finally, future word embeddings that use hyperbolic or elliptical geometries could be used to systematically capture nonlinear relations in language such as hierarchy or clustering. We argue that a wide range of techniques for productively developing and applying word embedding models to social and cultural inquiry are possible but yet to be developed. Nevertheless, Euclidean word embeddings are designed to model and evaluate intersecting dimensions of culture in a way that maps onto a wide range of cultural theory.

    Here we also pilot an approach to tame the interpretation of "unsupervised" methods like word embedding models. With supervised statistical classifiers analysts must first sample data, then train a predictive model, and finally extend or generalize that model to the remaining unsampled and unobserved data. It is easy to selectively interpret complex, unsupervised models by noting salient topics, word linkages or proximities that confirm expectations. Here we demonstrate how we can systematically compare words and dimensions against all others to enable complete or systematically sampled interpretations. Such sampling still requires a commitment to an underlying conceptual dimension, but enables systematic comparison with all meaningful linguistic dimensions, inferable from an extensive set of antonym pairs.

    A current limitation for the application of word embeddings is that they require a relatively large body of text if the output vector space is to capture subtle and complex associations of greatest interest to analysts of culture. Previous studies indicate that analogy tests

can only be reliably solved when input text comprises several million words or more (Hill et al. 2014). Nevertheless, that word embeddings are tailored to efficiently utilize very large corpora is also a strength. The amount of digitized text available for sociological analysis is growing at a rapid and accelerating rate (Salganik 2017). While social scientists have widely recognized that these vast archives contain a tremendous cache of cultural information and therefore hold enormous potential for cultural analysis, scholars have remained limited by the available tools for analyzing large text. Neural word embeddings present a method for producing a rich model of semantic relationships from corpora too large for techniques such as topic modeling or semantic network analysis. Furthermore, when an adequate amount of text is available, contemporary word embedding approaches are able to distill much more semantic information and more complex semantic relations than previous methods of computational text analysis. It is also important to note that word embedding models' algorithms have received intensive attention in computer science and natural language processing communities, and are being continuously improved, with more recent algorithms demonstrating advances in their ability to successfully model semantic relations with smaller bodies of text by leveraging subword information (Bojanowski et al. 2016; Joulin et al. 2016). As word embedding models become more widely used in the social scientific community, we expect them to enable the application of new data to classic sociological questions, and the productive identification and discovery of novel social and cultural problems.

The findings we present here demonstrate that word embedding models are able to faithfully capture complex cultural associations and dimensions from large bodies of text to a degree unapproachable with previous methods. High-dimensional word embeddings conform with, but also extend structural and field theories of culture, and critique the sufficiency of low dimensional theories, such as strong class or affect control, to explain the full range of strategic moves or human expressions. Through the engineering process of simultaneously capturing all associations expressed in language, neural embedding models reveal a stable, high-dimensional geometry of culture, which, like a many-faceted crystal, amplifies the competing cultural framings that enable coordinated and spontaneous social action.

TABLES AND FIGURES

**Table 1.** List of word pairs averaged to construct cultural dimensions of gender, class, and race

| Gender | Class | Race[†] |
|---|---|---|
| man – woman | rich – poor | black – white |
| men – women | richer – poorer | blacks – whites |
| he – she | richest – poorest | Blacks – Whites |
| him – her | affluence – poverty | Black – White |
| his – her | affluent – impoverished | African – European |
| his – hers | expensive – inexpensive | African – Caucasian |
| boy – girl | luxury – cheap | |
| boys – girls | opulent – needy | |
| male – female | | |
| masculine – feminine | | |

[†]Google Ngrams text was reduced to lowercase in preprocessing for all analyses to increase the word count for rare words. Because of this, the list of word pairs for used for ecological validation of Google Ngrams is *black-white, blacks-whites, african-european,* and *african-caucasian*.

**Table 2.** Pearson correlations between survey estimates and word embedding estimates for gender, class, and race associations

| | Gender | Class | Race |
|---|---|---|---|
| Google News *word2vec* Embedding | 0.88 | 0.52 | 0.70 |
| Common Crawl *GloVe* Embedding | 0.90 | 0.40 | 0.42 |
| Wikipedia *fastText* Embedding | 0.87 | 0.51 | 0.49 |
| Google Ngrams *word2vec* Embedding[†] | 0.78 | 0.60 | 0.17 |

N=59, except [†]N=56 where three words measured in the survey did not occur frequently enough in the text to appear in the word embedding.

**Table 3.** Percentage of statistically significant (p<.01) survey differences correctly classified in Google News word embedding model.

| | Sports | Food | Music | Occupations | Vehicles | Clothes | Names | All domains |
|---|---|---|---|---|---|---|---|---|
| Gender | 87.9% | 88.2% | 73.7% | 91.7% | 82.4% | 72.7% | 95.2% | 84.2% |
| Class | 96.4% | 94.1% | 100% | 53.2% | 94.4% | 92.5% | 86.4% | 71.2% |
| Race | 77.4% | 64.7% | 94.4% | 36.1% | 90.9% | 51.8% | 100% | 63.7% |

**Table 4.** Cultural dimensions with highest cosine similarity to gender and class dimensions in embedding models from United States and United Kingdom, 1890-1910.

| United States | | United Kingdom | |
|---|---|---|---|
| **Gender dimension nearest neighbors** | | | |
| 1. rugged–delicate | .219 | 1. tender–tough | -.211 |
|  | (.213, .224) |  | (-.221, -.202) |
| 2. soft–loud | -.209 | 2. timid–bold | -.191 |
|  | (-.216, -.201) |  | (-.200 -.182) |
| 3. tender–tough | -.202 | 3. rugged–delicate | 0.184 |
|  | (-.210, -.197) |  | (.176, .194) |
| 4. timid–bold | -.181 | 4. soft–loud | -.177 |
|  | (-.186, -.174) |  | (-.185, -.167) |
| 5. soft–hard | -.161 | 5. sweet–dry | -.172 |
|  | (-.168, -.158) |  | (-.176, -.166) |
| **Class dimension nearest neighbors** | | | |
| 1. weak-strong | -.292 | 1. bare_covered | -.255 |
|  | (-.301, -.287) |  | (-.263, -.244) |
| 2. fortunate-unfortunate | .291 | 2. employed_unemployed | .237 |
|  | (.286, .297) |  | (.231, .245) |
| 3. unhappy-happy | -.259 | 3. weak_strong | -.237 |
|  | (-.266, -.254) |  | (-.246, -.231) |
| 4. beautiful-ugly | .242 | 4. fortunate_unfortunate | .233 |
|  | (.238, .245) |  | (.224, .241) |
| 5. potent_impotent | .234 | 5. powerful_powerless | .226 |
|  | (.227, .244) |  | (.222, .230) |

Note: Reported cosine similarity is the mean cosine similarity of 20 subsamples. Bootstrapped 90% confidence intervals are given in parentheses.

**Figure 1**. Schematic figure that illustrates the structure of the descriptive problem Latent Semantic Analysis and neural word embeddings attempt to solve—how to represent all words from a corpus within a *k*-dimensional space that best preserves distances between words in local contexts. The solution, which we illustrate in subsequent figures, is a *n*-by-*k* matrix of values, bolded above for the example *k* = 3. These values can be understood as the coordinate for each word in a *k* dimensional space. In a real example, the matrix would be much more sparse and the word-by-*k* and *k*-by-context matrices would possess substantial symmetry. For a corpus of natural language texts, analysts have found that $k \geq 300$ is critical to minimize the error of this reconstruction required for solving analogy problems, which imply the preservation of underlying semantic dimensions.

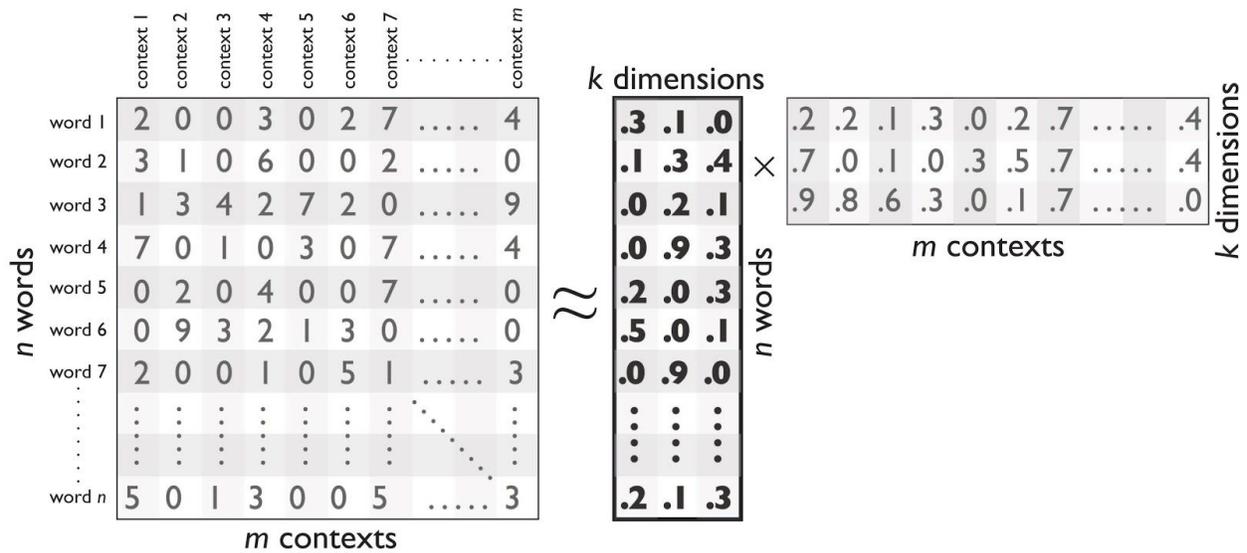

**Figure 2.** Conceptual diagram of (A) the construction of a cultural dimensions; (B) the projection of words onto the cultural dimension; and (C) the simultaneous projection of words onto multiple cultural dimensions.

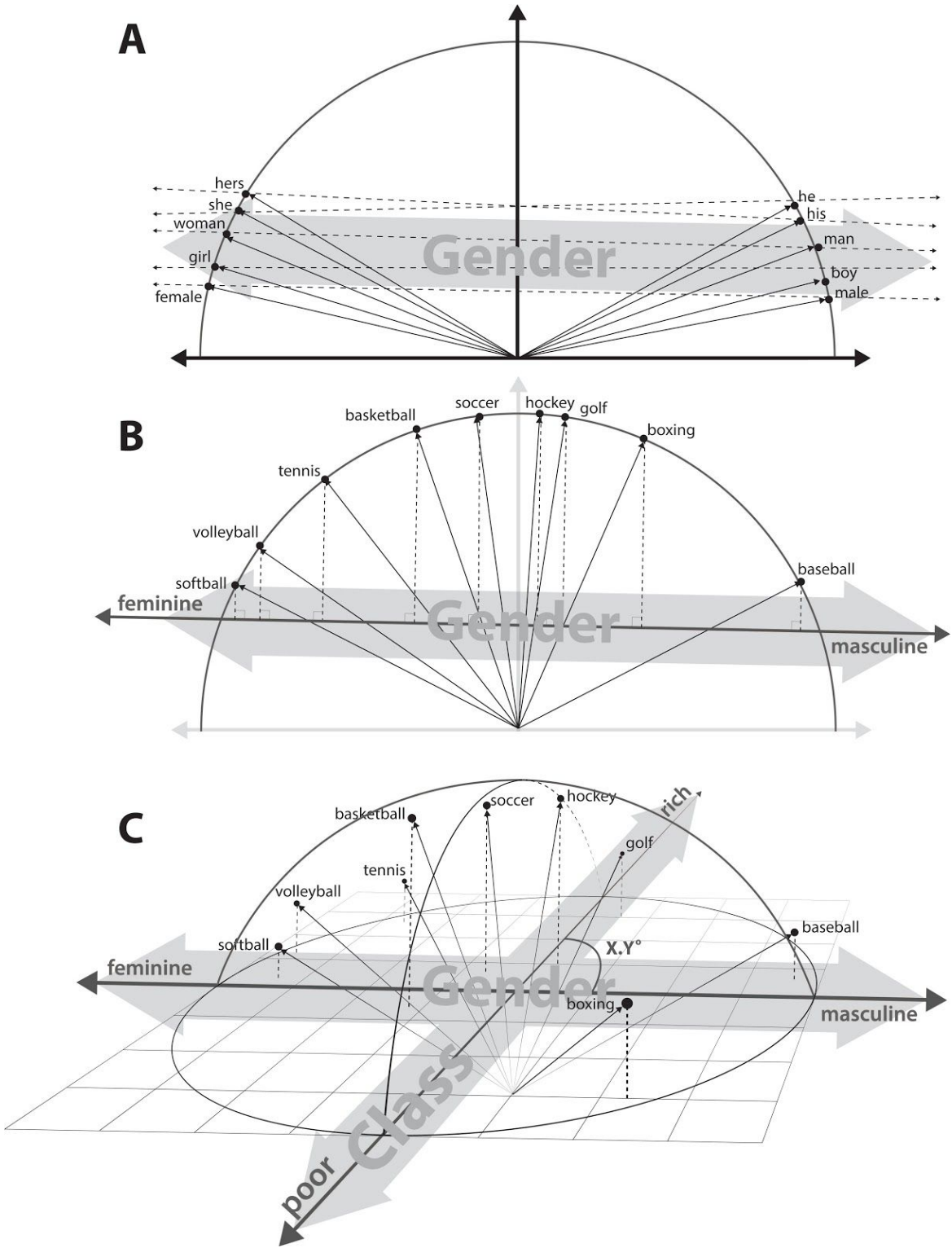

**Figure 3.** Projection of music genres onto race and class dimensions of the Google News word embeddings (grey) and average survey ratings of race and class associations (black).

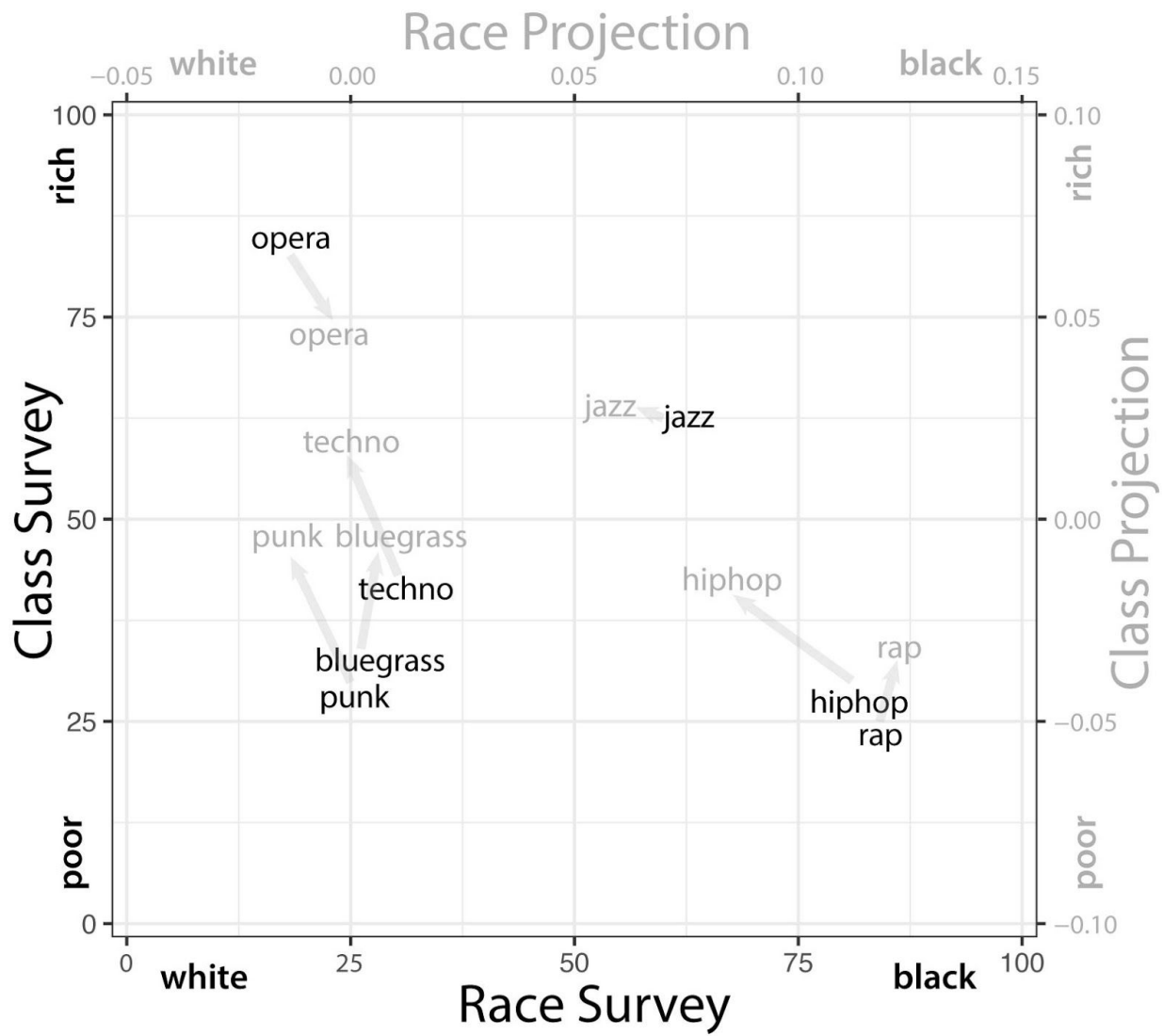

**Figure 4.** Projection of lifestyle-politics terms onto the dimensions of class and political ideology, 2015 Google News embedding.

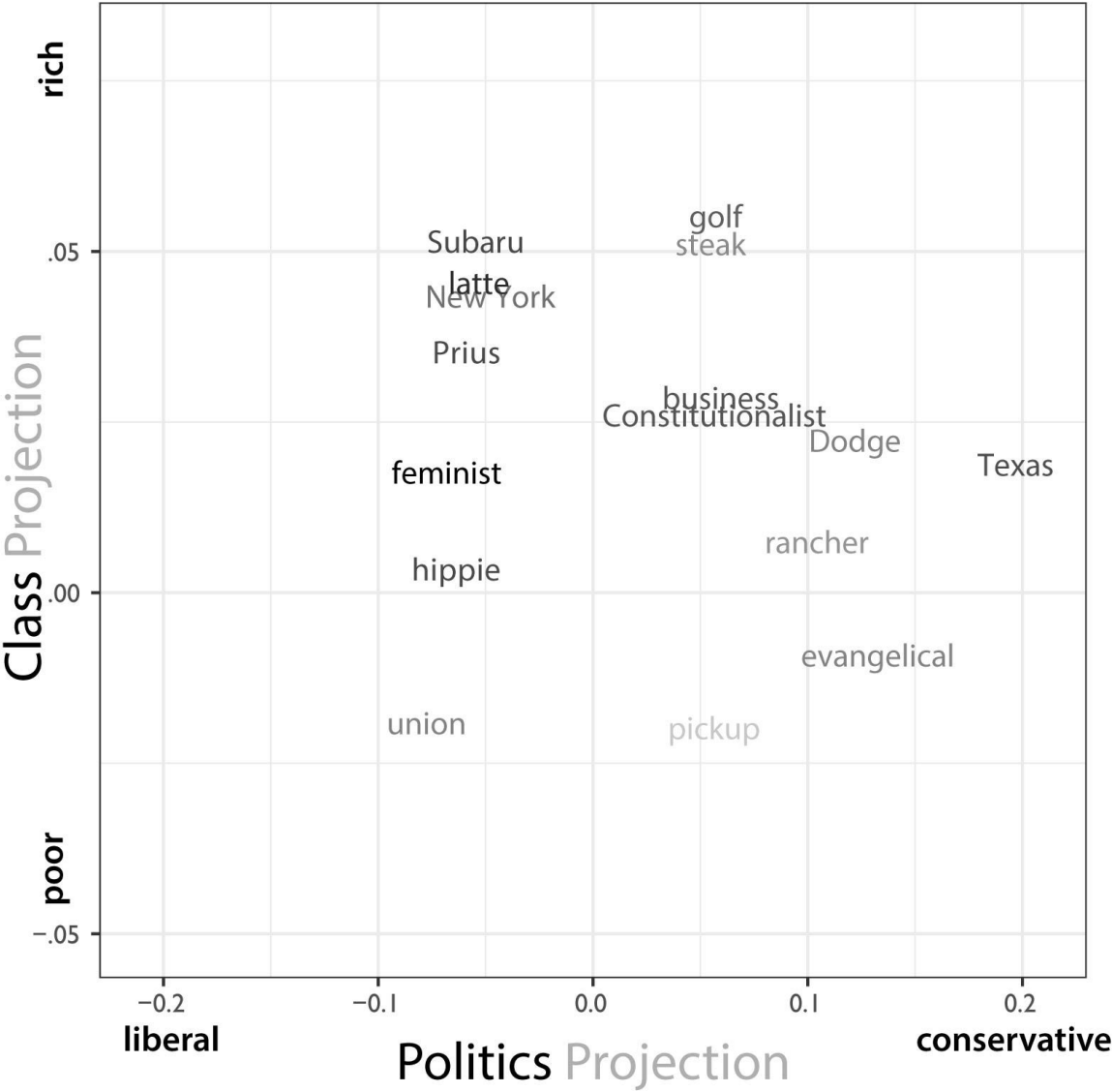

**Figure 5.** Cosine projection of occupations "journalist", "engineer", and "nurse" onto gender and class dimensions for each decade of the 20[th] century, with bootstrapped 90% confidence intervals.

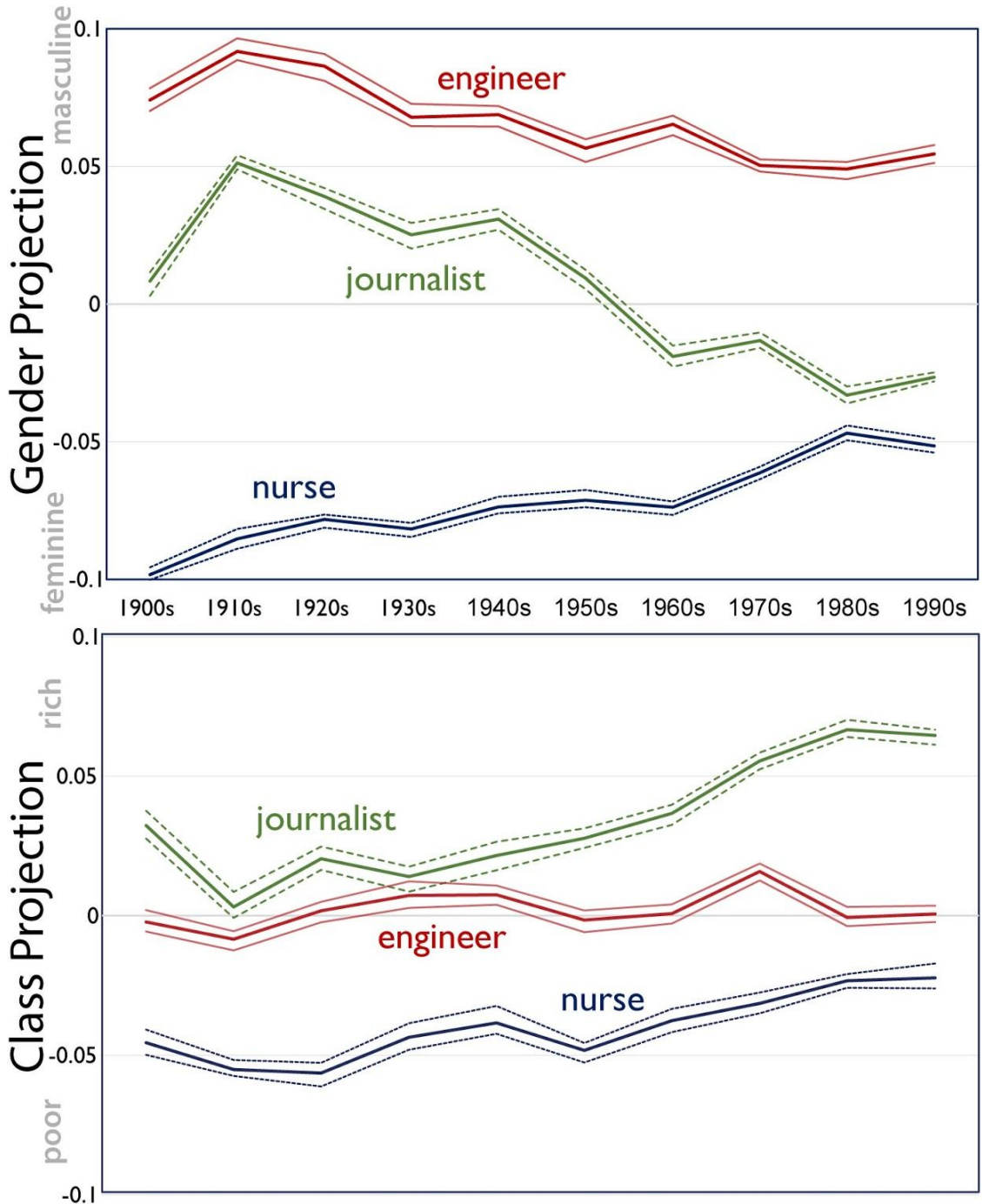

**Figure 6.** Cosine similarity between gender and class dimensions in each decade of the 20th century, with bootstrapped 90% confidence intervals.

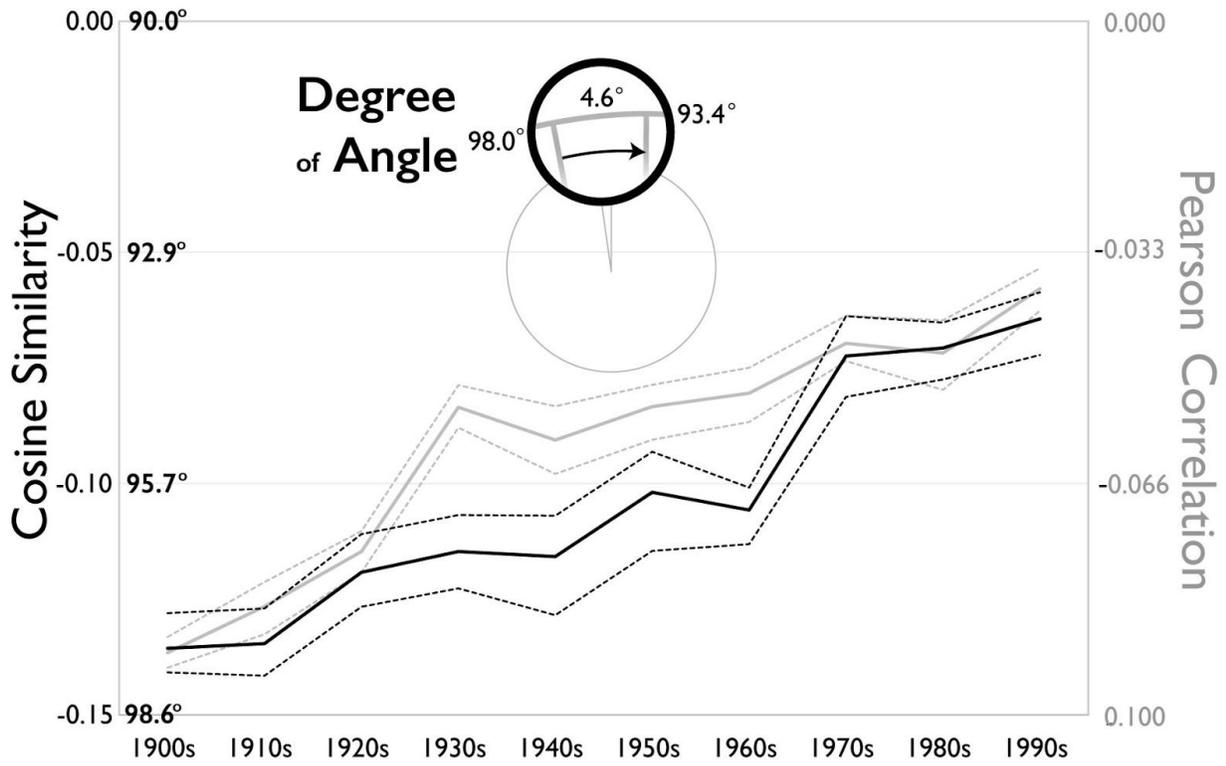

**Figure 7.** Cosine similarity of five cultural dimensions to the class dimension in text published in the United States for each decade of the 20th century, with bootstrapped 90% confidence intervals.

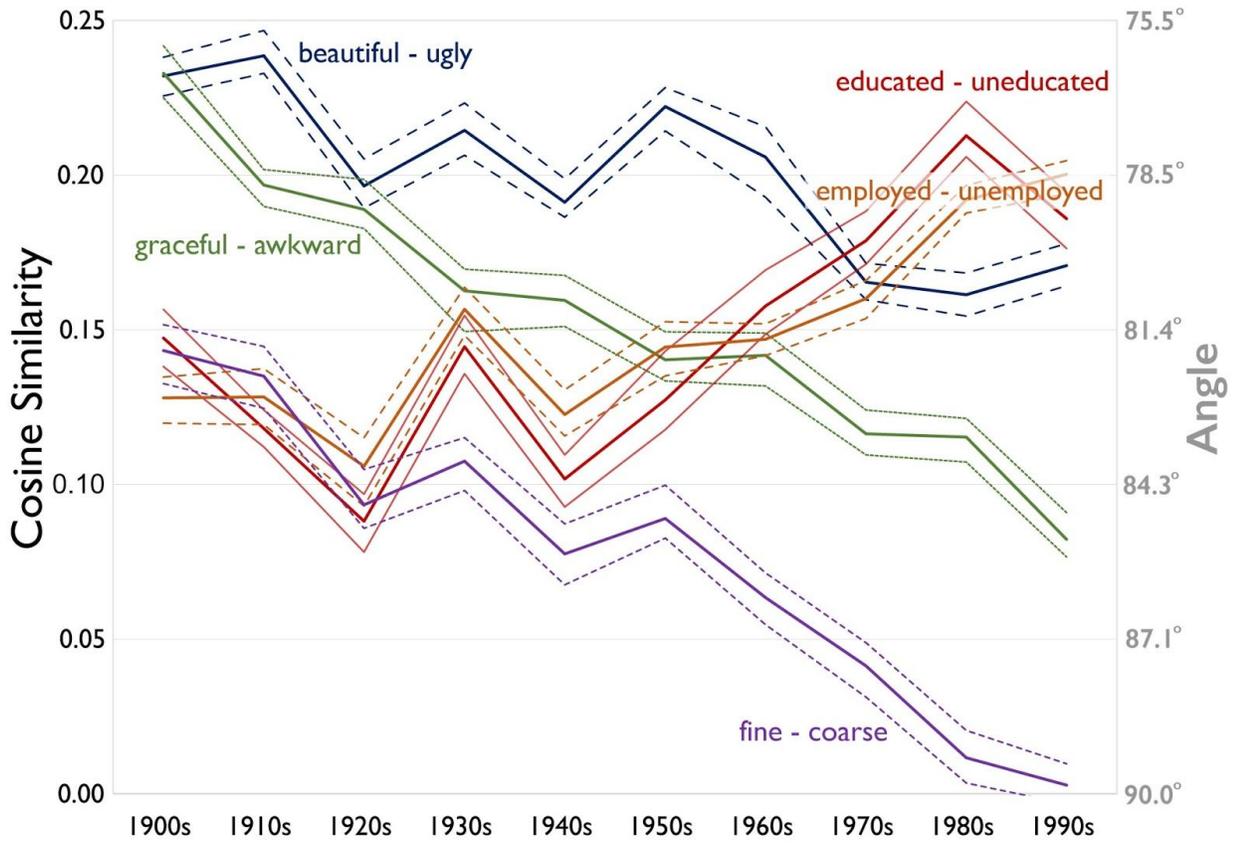

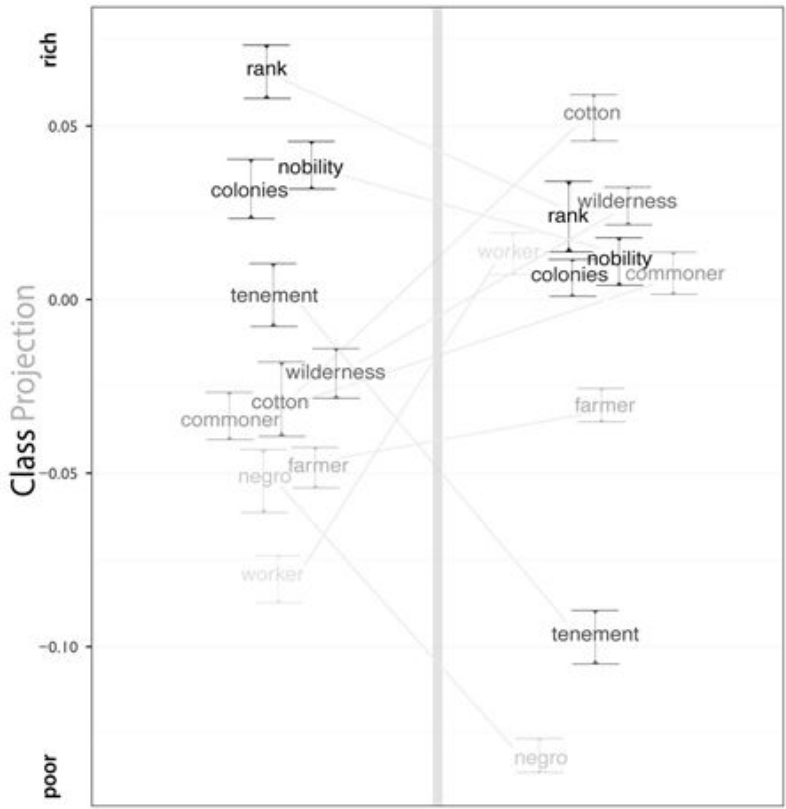
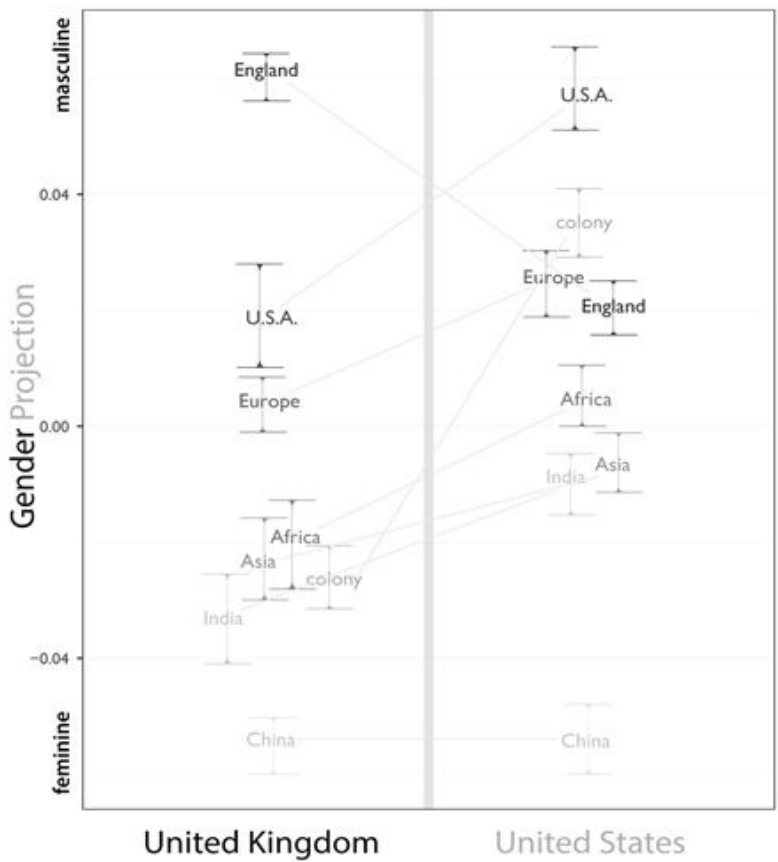

**Figure 8.** Comparison of gender and class projections from word embeddings trained on Google Ngrams from the United Kingdom and United States, 1890-1910. Text height represents the estimate and whiskers the 90% confidence intervals. Shading is proportional to alignment on the U.K. projection.

# APPENDIX A: SURVEY OF CULTURAL ASSOCIATIONS

**Table A1.** List of words included in cultural associations survey

| Occupations | Clothing | Sports | Music genres | Vehicles | Food | First names |
|---|---|---|---|---|---|---|
| Banker | Blouse | Baseball | Bluegrass | Bicycle | Beer | Aaliyah[†] |
| Carpenter | Briefcase | Basketball | Hiphop[†] | Limousine | Cheesecake | Amy |
| Doctor | Dress | Boxing | Jazz | Minivan | Hamburger | Connor |
| Engineer | Necklace | Golf | Opera | Motorcycle | Pastry | Jake |
| Hairdresser | Pants | Hockey | Punk | Skateboard | Salad | Jamal |
| Journalist | Shirt | Soccer | Rap | SUV | Steak | Molly |
| Lawyer | Shorts | Softball | Techno | Truck | | Shanice[†] |
| Nanny | Socks | Tennis | | | | |
| Nurse | Suit | Volleyball | | | | |
| Plumber | Tuxedo | | | | | |
| Scientist | | | | | | |

[†] Words that did not appear frequently enough in the 2000-2012 Google Ngrams to appear in the word embedding model and are therefore excluded from 2000-2012 Google Ngrams analyses.

This appendix details the Survey of Cultural Associations we fielded to provide a set of human-rated cultural evaluations against which to compare the results of the word embedding models. The survey was fielded through Amazon Mechanical Turk, an online service through which "requesters" can post a task and workers find and select tasks to complete in exchange for monetary compensation. Our survey was listed as "Sociological Survey" with the description "A fifteen-minute survey of cultural associations," with a compensation of $1.75. The task was only available to Mechanical Turk workers located in the United States. The survey was fielded in two waves, October 2016 and December 2017 to samples of 206 in 200 respectively, of which a total of 398 respondents completed the survey. The two waves are pooled in the analyses presented in this paper.

     A number of previous studies have found that Mechanical Turk surveys fare well when compared to surveys with probability sampling, particularly researchers measure and account for the socio-demographic characteristics of the sample (Levay, Freese, and Druckman 2016). Although Mechanical Turk's population of workers cannot be said to represent the general US population, it is characterized by considerable diversity along racial, gender, and socioeconomic lines (Huff and Tingley 2015).

     To mitigate any bias in estimates due to disproportionate representation of socio-demographic groups in the sample, we use post-stratification weighting to make our sample match the US general population. We took population estimates from the US Census Current Population Survey (CPS) of 2017 as population estimates for weighting our sample (United States Census Bureau 2017). We weighted along three strata: sex, education, and race. Sex is treated as two categories - male and female, education is divided into two categories - bachelor's degree or less than bachelor's degree, and race is divided into three strata: white, African American, or other. Results presented in this paper include post-stratification

weighting; however, additional analyses available upon request confirm that the inclusion of weights does not substantively alter results.

Table A2 displays basic demographic characteristics of the sample.

**Table A2.** Descriptive statistics of demographic makeup of Mechanical Turk sample and Census CPS sample.

|  | Mechanical Turk | Census CPS |
|---|---|---|
| Gender (1=female) | 43.47% | 51.76% |
| Education |  |  |
|   High school, GED, or less | 12.31% | 39.99% |
|   Some college | 26.88% | 18.83% |
|   Associate's degree | 10.05% | 9.75% |
|   Bachelor's degree | 43.47% | 20.03% |
|   Graduate degree | 7.29% | 11.39% |
| Race/Ethnicity |  |  |
|   African American | 6.53% | 12.52% |
|   White | 79.15% | 78.22% |
|   Other | 14.32% | 9.26% |
|   Hispanic | 9.82% | 15.92% |
| Age (mean) | 34.40 | 47.20 |
| N | 398 | 135,137 |

APPENDIX B: SUPPLEMENTARY FIGURES

**Figure B1.** Values for U.S. presidents, vice presidents, and presidential candidates projected onto the liberal-conservative dimension of the Google News word embedding.

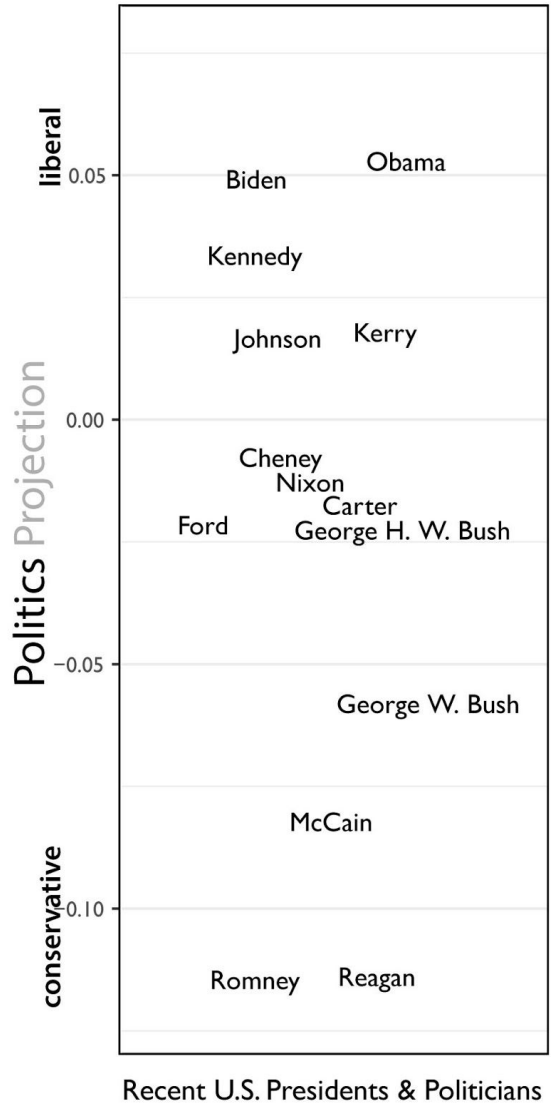

The model presented in Figure B1 overwhelmingly categorizes politicians correctly; that is, Democrats project at values greater than 1 and Republicans project less than 1. The one exception is Jimmy Carter, who is notable for his personal religious conservatism. To obtain the most accurate possible estimates, projections for each term were calculated by averaging the projections of several relevant word vectors. The Google News embedding preserves commonly occuring sets of two or three words as vectors, such as "John_F._Kennedy." In all cases, vectors for the last name (e.g. "Obama") and the full name ("Barack_Obama") were averaged. In the cases of John F. Kennedy and Lyndon Johnson, the nicknames "J.F.K." and "L.B.J." were also included in the calculation of the averaged vectors. Following this method, it is possible to obtain more robust vectors for single entities, such as a person, as well as for the categories themselves, such as liberal-conservative, by averaging multiple synonymous terms.

**Figure B2.** Popular boys' and girls' names projected on the gender dimension of word embedding models of Google Ngram texts from each decade of the 20[th] century.

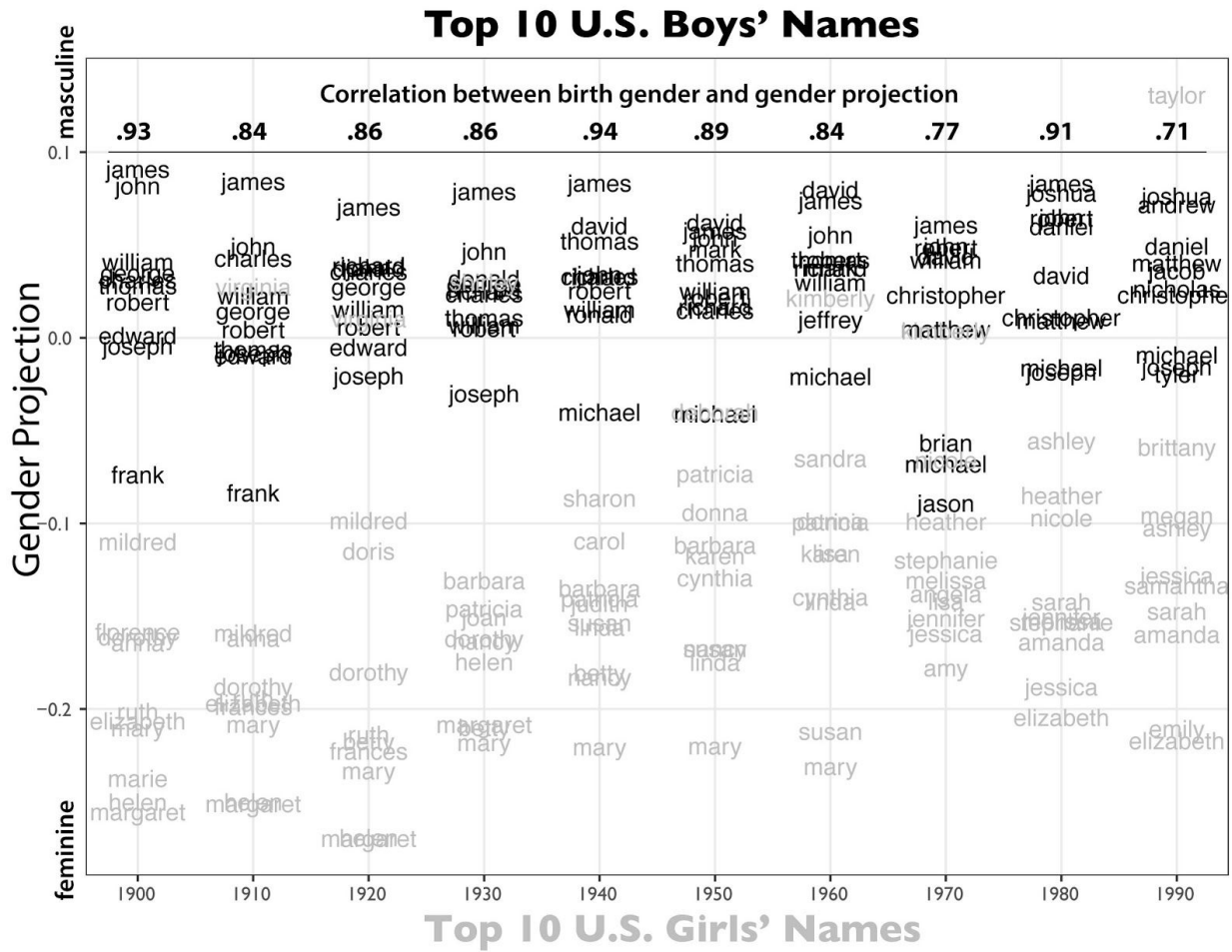

The top ten most popular names for newborn boys and girls each decade were retrieved from the Social Security Record. We then calculated the projection of each name onto the gender dimension of the word embedding trained on the Google Ngrams corpus for the corresponding decade. We induced a lag of two decades between the decade of name popularity and text used (e.g. names popular in the 1880s are projected with 1900s text) because many names did not appear frequently enough in the corpus in the decade of their popularity to be represented as a vector. It is not entirely surprising that individuals with a given name will not appear frequently in texts until they have attained adulthood, so we may think of these as popular young adult names, rather than popular infant names for the respective decade. Boys' names are depicted in black and girls' names in grey. We see a near perfect sorting, with the few exceptions including unisex names like "Taylor" and "Virginia", which would commonly appear in texts with reference to the state rather than person. Overall, the persistence of high correlation between gender projection and the sex associated with the name suggests that the gender dimension remains a useful indicator for the historic texts.